\documentclass[acmtog,screen]{acmart}

\acmSubmissionID{2215}

\copyrightyear{2026}
\acmYear{2026}
\setcopyright{cc}
\setcctype{by}
\acmConference[SA Conference Papers '26]
  {SIGGRAPH Asia 2026 Conference Papers}
  {December 01--04, 2026}
  {Kuala Lumpur, Malaysia}
\acmBooktitle{SIGGRAPH Asia 2026 Conference Papers
  (SA Conference Papers '26), December 01--04, 2026,
  Kuala Lumpur, Malaysia}
\acmDOI{10.1145/3829340.3842330}
\acmISBN{979-8-4007-2842-6/2026/12}

\usepackage{xspace}
\usepackage{graphicx}
\usepackage{booktabs}
\usepackage{xcolor}
\usepackage{colortbl}
\usepackage{multirow}
\usepackage{pifont}
\usepackage{microtype}
\usepackage{url}
\usepackage{hyperref}
\usepackage[table]{xcolor}

\usepackage[capitalize]{cleveref}
\crefname{section}{Sec.}{Secs.}
\Crefname{section}{Section}{Sections}
\Crefname{table}{Table}{Tables}
\crefname{table}{Tab.}{Tabs.}

\usepackage{caption}

\definecolor{bestgreen}{RGB}{153,200,76}
\definecolor{worstred}{RGB}{192,0,0}

\definecolor{cbad}{HTML}{FFD0D0}
\definecolor{cmedium}{HTML}{FFF0D0}
\definecolor{cgood}{HTML}{90C060}

\newlength\savewidth

\definecolor{GreenColor}{rgb}{0.137,0.573,0.565}
\definecolor{DeltaColor}{rgb}{0.039,0.73,0.71}
\definecolor{SigmaColor}{rgb}{0.98,0.45,0.0}
\definecolor{AlphaColor}{rgb}{0,0,0.8}
\definecolor{BetaColor}{rgb}{0.8,0,0.8}
\definecolor{GammaColor}{rgb}{0.514,0.34,0.224}
\definecolor{EpsilonColor}{rgb}{0.353,0.725,0.906}
\definecolor{PurpleColor}{HTML}{9839ff}
\definecolor{RedColor}{rgb}{0.949,0.275, 0.224}
\definecolor{citecolor}{HTML}{0071bc}

\definecolor{deepred}{HTML}{940000}
\definecolor{cvprblue}{rgb}{0.21,0.49,0.74}
\hypersetup{linkcolor=deepred,urlcolor=cvprblue,citecolor=[rgb]{0.4,0.15,0.95}}

\definecolor{markerDarkBlue}{HTML}{1B3A8C}
\definecolor{markerBlue}{HTML}{3B6AB8}
\definecolor{markerMidBlue}{HTML}{6FA8D6}
\definecolor{markerLightBlue}{HTML}{A8D5E5}
\definecolor{markerRed}{HTML}{E63946}
\definecolor{markerYellow}{HTML}{F4C842}
\newcommand{\fullcirc}{\raisebox{-0.05ex}{\scalebox{1.25}{$\bullet$}}}
\newcommand{\emptycirc}{\raisebox{0.0ex}{\scalebox{1.35}{$\circ$}}}

\newcommand{\etc}{\mbox{etc}\xspace}

\newcommand{\ie}{\mbox{i.e.}\xspace}

\newcommand{\modelname}{\mbox{DirtyMoCap}\xspace}
\newcommand{\longtitle}{Robust Motion Capture from Unconstrained Markers}
\newcommand{\ourtitle}{\textbf{\modelname}: \longtitle}

\newcommand{\suppl}{\textcolor{magenta}{\emph{Sup.Mat.}}\xspace}

\newcommand{\xmark}{\textcolor{RedColor}{\ding{55}}\xspace}
\newcommand{\cmark}{\textcolor{GreenColor}{\ding{51}}\xspace}

\begin{document}

\title{\ourtitle}

\author{Long Wang}
\affiliation{%
  \institution{Zhejiang University}
  \city{Hangzhou}
  \country{China}}
\affiliation{%
  \institution{Westlake University}
  \city{Hangzhou}
  \country{China}}
\email{wanglong@westlake.edu.cn}

\author{Shuting Zhao}
\affiliation{%
  \institution{Westlake University}
  \city{Hangzhou}
  \country{China}}
\affiliation{%
  \institution{Fudan University}
  \city{Shanghai}
  \country{China}}
\email{zhaoshuting@fudan.edu.cn}

\author{Shen Yan}
\affiliation{%
  \institution{National University of Defense Technology}
  \city{Changsha}
  \country{China}}
\email{yanshen12@nudt.edu.cn}

\author{Siyuan Yu}
\affiliation{%
  \institution{Westlake University}
  \city{Hangzhou}
  \country{China}}
\email{yusiyuan@westlake.edu.cn}

\author{Xiaoben Li}
\affiliation{%
  \institution{Zhejiang University}
  \city{Hangzhou}
  \country{China}}
\affiliation{%
  \institution{Westlake University}
  \city{Hangzhou}
  \country{China}}
\email{lixiaoben@westlake.edu.cn}

\author{Zeyu Cai}
\affiliation{%
  \institution{Westlake University}
  \city{Hangzhou}
  \country{China}}
\affiliation{%
  \institution{Nanjing University}
  \city{Nanjing}
  \country{China}}
\email{caizeyu010612@gmail.com}

\author{Yumeng Hou}
\affiliation{%
  \institution{National University of Singapore}
  \city{Singapore}
  \country{Singapore}}
\email{yumeng.hou@nus.edu.sg}

\author{Yuliang Xiu}
\authornote{Corresponding author.}
\affiliation{%
  \institution{Westlake University}
  \city{Hangzhou}
  \country{China}}
\email{xiuyuliang@westlake.edu.cn}

\begin{abstract}
Optical motion capture delivers high-fidelity human motion, but its reliance on strict marker layouts and clean trajectories severely limits its real-world applicability. In practice, tracking systems frequently output unconstrained markers—sparse, noisy, and unordered point clouds with unknown or varying configurations.
To bridge the gap between corrupted raw markers and parametric human models, we introduce \modelname, a robust, marker-layout-free framework. Our core insight is to map unordered marker observations to a fixed set of ``proxy anchors''—comprising skeletal joints and body surface points—acting as a stable intermediate representation. We first initialize and track these anchors over long sequences using a recurrent sliding-window architecture. Then, a custom differentiable Gauss-Newton solver fits the SMPL-H model to the tracked anchors to recover full-body pose, translation, and shape. By explicitly deriving geometric residuals, our solver learns adaptive observation confidence, smoothness, and prior weights end-to-end, adapting dynamically to the reliability of the input data.
Extensive experiments on diverse, noisy marker configurations demonstrate that \modelname successfully generalizes across arbitrary layouts using only a single trained model. It consistently outperforms state-of-the-art configuration-specific baselines in both joint and vertex reconstruction accuracy,
while our custom CUDA solver achieves up to a 100$\times$ speedup over standard PyTorch implementations. 
We further apply \modelname to heterogeneous raw optical MoCap recordings of traditional Chinese martial arts, yielding a Kung Fu motion dataset of temporally coherent SMPL-H reconstructions.
Code and data for this paper are at \href{https://wanglongzju.github.io/DirtyMoCap-Project-Page/}{project page}.

\end{abstract}

\begin{teaserfigure}
  \includegraphics[width=\textwidth]{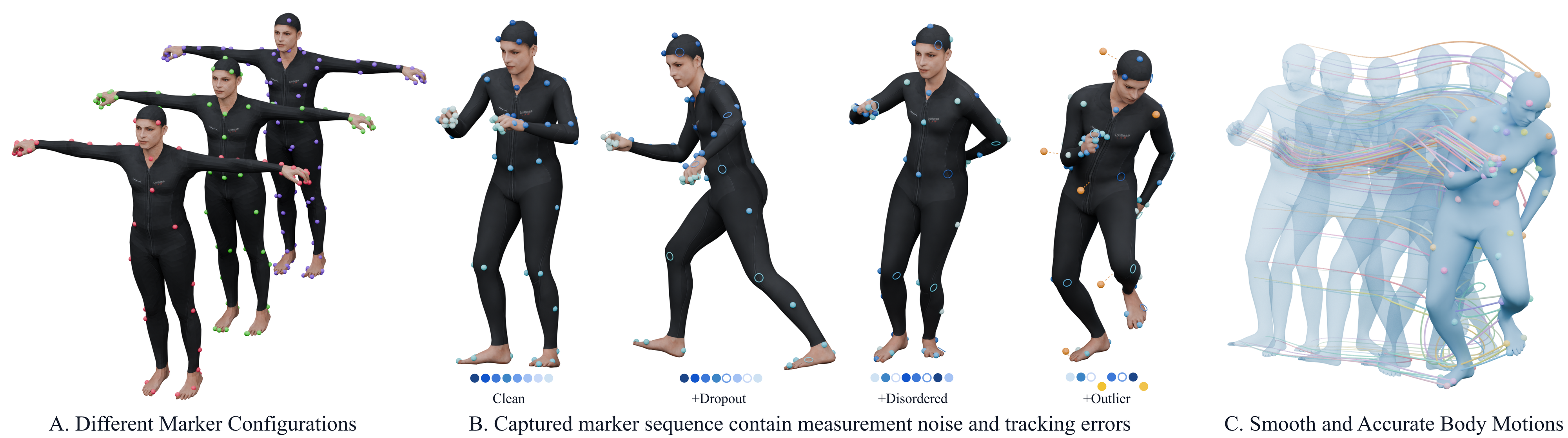}
  \caption{\textbf{\modelname robustly recovers human motion from dirty MoCap.} (\textbf{A}) MoCap systems vary in marker configuration, differing in layout and marker count. (\textbf{B}) Captured marker sequence may contain: 1) clean, correctly ordered markers \textcolor{markerDarkBlue}{\fullcirc}\,\textcolor{markerBlue}{\fullcirc}\,\textcolor{markerMidBlue}{\fullcirc}\,\textcolor{markerLightBlue}{\fullcirc}, 2) dropout markers due to occlusion \textcolor[HTML]{6D9EEB}{\emptycirc}\,\textcolor[HTML]{c9daf8}{\emptycirc}, 3) reordered markers upon reappearance after occlusion \textcolor{markerLightBlue}{\fullcirc}\,\textcolor{markerDarkBlue}{\fullcirc}\,\textcolor{markerMidBlue}{\fullcirc}\,\textcolor{markerBlue}{\fullcirc}, and 4) outlier markers \textcolor{markerYellow}{\fullcirc}\,\textcolor{markerYellow}{\fullcirc} from measurement noise or tracking errors. 
  (\textbf{C}) Despite these corruptions, \modelname recovers a smooth anchor trajectory, enabling stable geometric constraints for parametric body fitting.
  }
  \label{fig:taster}
\end{teaserfigure}



\maketitle

\section{Introduction}

Commercial optical motion-capture systems, such as Vicon, Qualisys and OptiTrack, are widely used to acquire high-fidelity human motion data for animation, biomechanics, sports science, and clinical analysis. Despite their high accuracy, these systems rely on carefully designed marker configurations and often require substantial manual effort for marker cleaning.

A marker configuration defines \textit{where} markers are placed on the body, which is typically represented by a parametric human model, such as SMPL-H~\cite{pavlakos2019expressive}, MHR~\cite{ferguson2025mhr}, and \etc. Commercial pipelines assume specific marker-set designs; without the correct configuration, fitting a body model to captured markers is difficult. Even when the configuration is known, placement errors and occlusion-induced ``ghost'' points can corrupt tracking and degrade downstream skeletal fitting. Markers that reappear after occlusion may also receive new labels, breaking temporal consistency. In practice, optical MoCap faces four key challenges: 
1) unknown or varying marker configurations (\cref{fig:taster}-a); and corrupted markers that are 2) dropout/occluded, 3) disordered/shuffled, or 4) outlier/jitter-corrupted markers (\cref{fig:taster}-b). We refer to these conditions as ``unconstrained markers,'' in contrast to constrained markers with known layouts, consistent ordering, and clean captures.

These challenges motivate a configuration-agnostic and noise-robust optical MoCap system. MoCap-Solver~\cite{chen2021mocap} predicts clean marker trajectories and skeletal motion using representations of the skeleton, marker configuration, and pose-dependent marker reliability; SOMA~\cite{ghorbani2021soma} labels unordered observations while rejecting spurious points; and LocalMoCap and RoMo~\cite{pan2023locality,pan2024romo} exploit marker--joint graphs and spatial locality to recover occluded markers, detect tracking errors, and reconstruct body and hand motion. However, these methods remain configuration-specific, relying on predefined marker sets, skeletal associations, or training-time neighborhood structures. DAMO~\cite{kim2024damo} supports arbitrary configurations, but directly regressing parametric body parameters from sparse or noisy markers can limit accuracy.

To address these limitations, we propose \modelname, which handles noisy and unordered markers under arbitrary configurations by predicting a structured intermediate representation: proxy anchors (``anchors''). Anchors are defined by combining selected skeletal joints with surface vertices on a parametric human model. Unlike marker observations, whose configurations and identities may be unknown, anchor identities are fixed by design and remain consistent across body models. This removes the need to specify marker configurations during training or inference while providing stable geometric constraints for temporal tracking and parametric body fitting. 
Inspired by long-range point tracking~\cite{karaev2024cotracker,xiao2024spatialtracker}, we estimate anchor trajectories through recurrent sliding-window refinement, where frame-wise marker--anchor attention aggregates observation evidence and anchor-wise temporal attention propagates it across frames through ordered anchors.

Given predicted anchor trajectories, the fixed anchor-to-body correspondences provide direct geometric constraints for fitting a parametric human model (\ie, SMPL-H) and recovering temporally coherent pose, global translation, and shape. This fitting stage is widely adopted as a post-optimization step in human-centric tasks, including MoCap data standardization~\cite{AMASS:ICCV:2019,ghorbani2021soma} and human mesh recovery~\cite{patel2025camerahmr, li2025etch, cai2026omnifit}. However, conventional post-optimization relies on hand-crafted weights and empirical hyperparameters that remain fixed across all inputs, limiting its ability to adapt to varying correspondence reliability or prior strength.

Recent advances in camera pose~\cite{teed2021droid, sarlin2021back} and object pose~\cite{chen2022epro, wang2023deep} estimation make optimization differentiable by explicitly deriving geometric residuals, enabling end-to-end learning of solver parameters and weighting policies. Extending this idea to articulated human motion is more challenging because of the high-dimensional kinematic chain, pose-dependent deformations, Linear Blend Skinning, and shared shape parameters.
To address this, we formulate anchor-based body fitting as a differentiable Gauss--Newton solver, so supervision on optimized body parameters can back-propagate through the optimization process itself. This lets the model learn input-dependent correspondence confidence, smoothness weights, and prior weights without direct supervision on these terms, improving both accuracy and robustness. We further implement efficient CUDA kernels for derivatives along the human kinematic chain, achieving up to 100\(\times\) speedup over a PyTorch implementation for efficient training and inference.

We further apply \modelname to a heterogeneous collection of raw optical MoCap recordings of traditional Chinese martial arts, for which marker configurations and marker identities are unavailable. The collection contains 134 real motion sequences, averaging approximately 8,500 frames per sequence and totaling about 180 minutes across 22 normalized martial-arts style labels. Processing these noisy and unordered observations with \modelname yields temporally coherent SMPL-H motion and enables the construction of the Kung Fu motion dataset described in \cref{sec:dataset}.
Our main contributions are summarized as follows:

\begin{itemize}
    \item \textbf{Proxy Anchors.} We introduce an intermediate representation that maps noisy, unordered, and incomplete optical markers to a fixed set of body anchors. Anchors remain directly comparable to marker observations while preserving stable correspondences to the parametric body model, removing the need for predefined layouts or explicit labeling.
    \item \textbf{Marker-to-Anchor.} We propose a recurrent sliding-window tracker that estimates anchor trajectories via frame-wise marker--anchor interaction and anchor-wise temporal propagation, enabling robust and stable long-sequence tracking.
    \item \textbf{Anchor-to-Body.} We develop a differentiable Gauss--Newton solver that recovers pose, translation, and shape from tracked anchors.The solver learns adaptive observation, smoothness, and prior weights end-to-end to improve robustness, while our CUDA implementation provides up to 100$\times$ speedup over a standard PyTorch version.
    \item \textbf{HKMALA-Motion Dataset.} We apply \modelname to noisy, unordered raw markers with unknown marker configurations and marker identities, recovering temporally coherent SMPL-H motion to construct a new Kung Fu motion dataset. 
\end{itemize}

\section{Related Work}
\noindent\textbf{Marker-based Motion Capture.}
Optical marker-based motion capture acquires accurate 3D human motion, but raw trajectories require skeletal solving, marker-dropout recovery, and denoising. Traditional methods use articulated skeletal fitting~\cite{kirk2004skeletal}, robust matrix completion~\cite{feng2014exploiting}, low-dimensional Kalman smoothing~\cite{burke2016estimating}, soft skeleton constraints with model averaging~\cite{tits2018robust}, and motion self-similarity analysis~\cite{aristidou2018self}; they generally assume labeled trajectories, fixed marker identities, or known marker--body configurations.
Learning-based methods automate marker cleaning, labeling, and motion solving. Holden~et al.~\cite{holden2018robust} directly estimate joint transformations from denoised marker observations; MoCap-Solver~\cite{chen2021mocap} jointly models motion, marker configuration, and a template skeleton to recover clean marker trajectories and skeletal motion within a known configuration family; and SOMA~\cite{ghorbani2021soma} labels unordered point clouds using a transformer and optimal transport while rejecting spurious points. More recent methods exploit marker structure and temporal information: LocalMoCap~\cite{pan2023locality} models markers and joints as heterogeneous graph nodes for marker completion, tracking-error detection, and body-and-hand motion solving; RoMo~\cite{pan2024romo} decomposes full-body unlabeled MoCap into alignment, body-part segmentation, and part-specific labeling with joint positions as an intermediate representation; DAMO~\cite{kim2024damo} learns a deep solver for arbitrary marker configurations; and OpenMoCap~\cite{qian2025openmocap} introduces marker--joint chain inference for severe real-world marker-dropout patterns. Although these methods improve robustness to noise, occlusion, and configuration variation, most still rely on marker labels, marker vocabularies, local neighborhood structures, or explicit configuration inference. In contrast, our method uses geometric anchors as configuration-independent intermediate correspondences without direct marker labeling or configuration inference.

\noindent\textbf{Differentiable Solver.}
Differentiable optimization has been widely explored in camera and object pose estimation, where classical geometric solvers are embedded into neural pipelines and trained end-to-end. In camera pose estimation and SLAM, PixLoc~\cite{sarlin2021back} refines camera poses by feature--metric alignment, while DROID-SLAM~\cite{teed2021droid}, DROID-W~\cite{li2026droid}, and MegaSaM~\cite{li2025megasam} incorporate differentiable bundle adjustment or learned optimization for robust SfM estimation. Similar ideas have also been applied to object pose estimation and tracking: RePose~\cite{iwase2021repose} and RNNPose~\cite{xu2022rnnpose} use differentiable Levenberg--Marquardt refinement, EPro-PnP~\cite{chen2022epro} formulates PnP as a probabilistic differentiable layer, and DeepAC~\cite{wang2023deep} integrates learned contour alignment with pose optimization. These methods demonstrate that differentiable solvers can learn task-specific correspondences, weights, or update strategies from final pose supervision. However, human motion recovery involves a high-dimensional articulated kinematic chain, pose-dependent deformation, Linear Blend Skinning, shared shape parameters, and temporal constraints. To handle these challenges, our method extends differentiable optimization by introducing a Gauss--Newton solver that optimizes human pose, translation, and shape from tracked anchors while learning anchor confidence, and adaptive weights of  smoothness and prior end-to-end.

\section{Method}

\begin{figure*}[t]
    \includegraphics[width=0.92\textwidth]{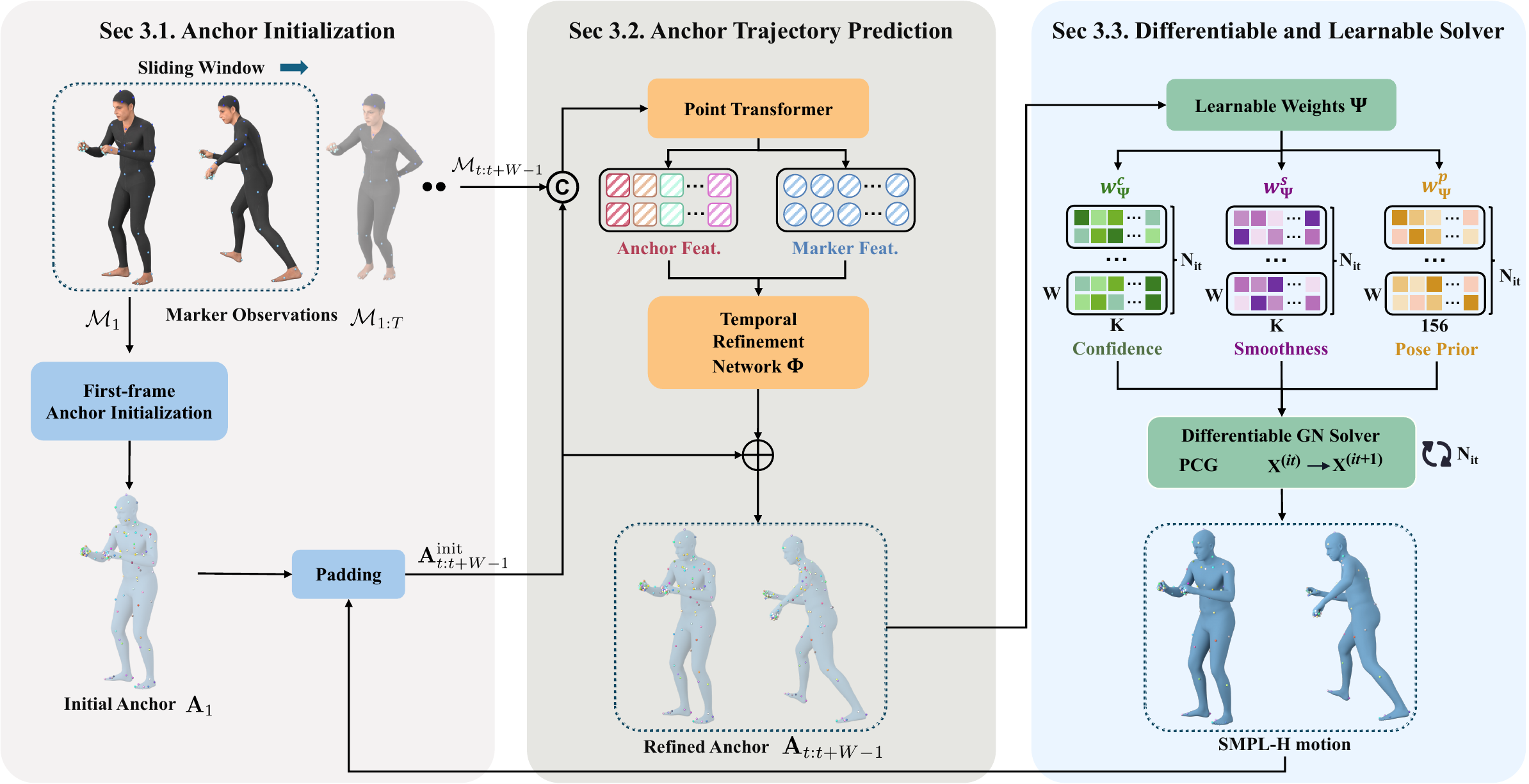}
    \caption{\textbf{Overview of \modelname}. 
    The Anchor Initialization step (\cref{sub:anchor_initialization}) uses the first-frame markers $\mathcal{M}_{1}$ to estimate initial anchors $\mathbf{A}_1$ and replicates them across the remaining frames in the current window to form $\mathbf{A}^{\mathrm{init}}_{t:t+W-1}$. The resulting anchor sequence, together with $\mathcal{M}_{t:t+W-1}$, is then passed to the Anchor Trajectory Prediction module (\cref{sub:anchor_tracking}) to refine $\mathbf{A}_{t:t+W-1}$. Finally, a differentiable learnable Gauss--Newton solver (\cref{sub:differentiable_solver}) optimizes the SMPL-H parameters, including pose, global translation, and shape. The estimates from the current window initialize the next one.}
    \label{fig:main_overview}
\end{figure*}

Given a sequence of unordered and noisy marker observations, $\mathcal{M}_{1:T}$, our goal is to reconstruct smooth human motion parameterized by SMPL-H~\cite{pavlakos2019expressive}. Specifically, we estimate pose $\theta_{1:T}$, global translation $\tau_{1:T}$, and shape $\beta$.
In contrast to prior optical MoCap solvers~\cite{ghorbani2021soma, pan2023locality, pan2024romo, qian2025openmocap} that assume a predefined marker configuration or consistent marker identities across frames, our method does not require a known marker configuration of the input markers.
To bridge the gap between unordered marker observations and the parametric body model, we reconstruct an intermediate set of geometric anchors, $\mathbf{A}_{1:T}$. These anchors are defined in Euclidean space and are therefore directly comparable to the marker observations, while also maintaining fixed correspondences to the SMPL-H model. Specifically, the anchor set consists of SMPL-H skeletal joints $\mathbf{J}_{1:T}$ and a set of predefined surface anchors $\mathbf{S}_{1:T}$ attached to mesh vertices.
As illustrated in ~\cref{fig:main_overview}, our method proceeds in three stages. First, we initialize the full anchor set in the first frame (~\cref{sub:anchor_initialization}). We then track the anchors over time to obtain a temporally coherent anchor trajectory (~\cref{sub:anchor_tracking}). Finally, given the established anchor-to-body correspondences, we solve for the SMPL-H parameters using a differentiable and learnable Gauss--Newton solver (~\cref{sub:differentiable_solver}).

\subsection{Anchor Initialization}
\label{sub:anchor_initialization}

The first stage of our pipeline initializes the full anchor set $\mathbf{A}_{1}$ in the first frame from marker observations. This is challenging because the input markers are unordered, noisy, and do not follow a known marker configuration. Directly predicting all anchors from such observations is highly ambiguous, especially in the presence of large global motion or partial observations. To address this, we adopt a coarse-to-fine initialization strategy. We first estimate a small set of stable anatomical anchors to establish a body-centric canonical frame, and then predict the full anchor set in this canonical space.

\begin{figure}[t] \centering
  \includegraphics[width=0.48\textwidth]{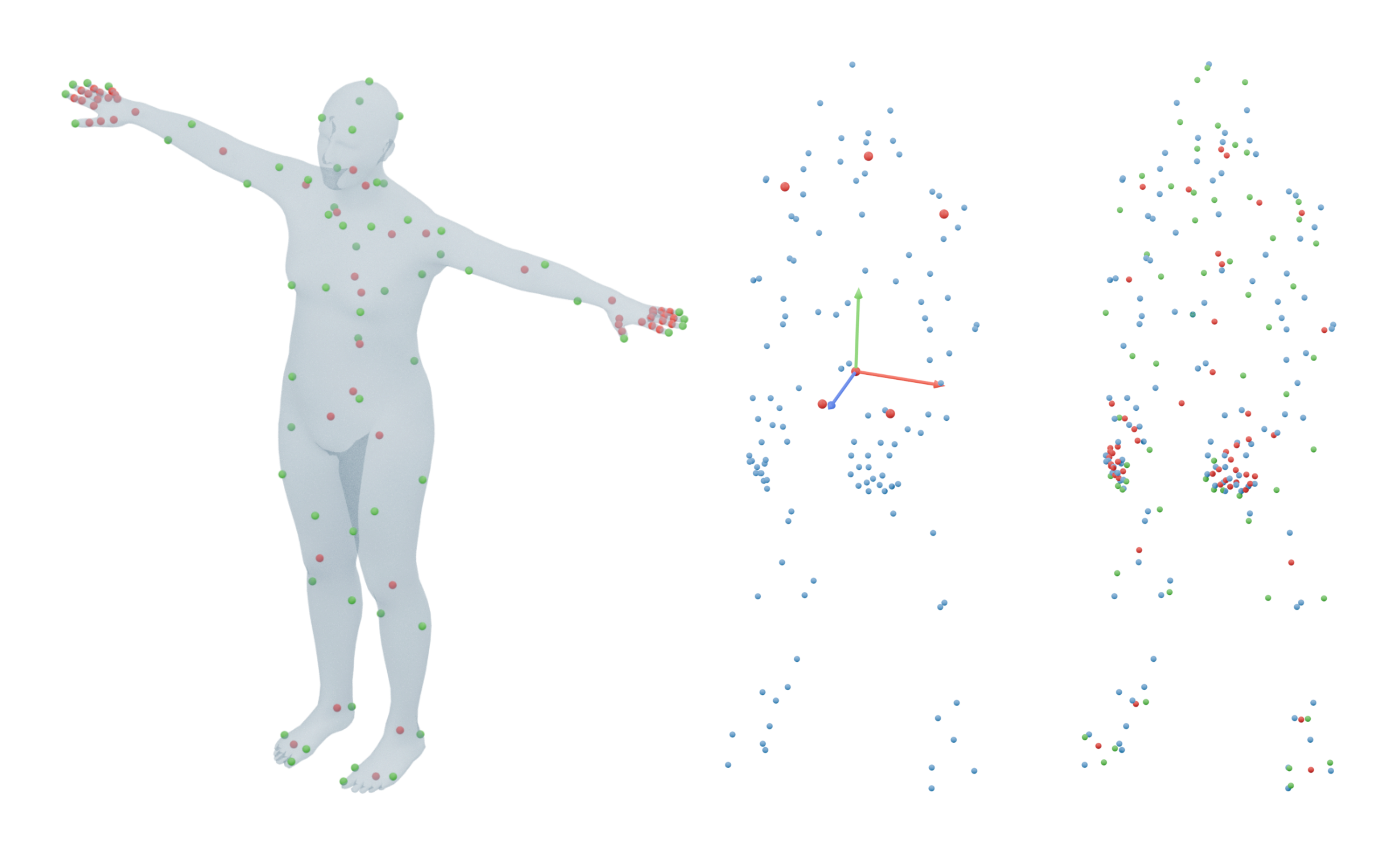}
  \caption{\textbf{Anchor representation.}
    Left: The anchors are shown on the template SMPL-H mesh and consist of 52 \textcolor{red}{joints} and 61 \textcolor{green!60!black}{surface anchors}. Middle:  \textcolor[HTML]{6F95C5}{Markers} are used in the coarse alignment stage to predict 6 \textcolor{red}{anatomical anchors} for canonicalization. Right: The full anchor set obtained from the Anchor Initialization.}
  \label{fig:body_centric}
\end{figure}

Given the marker observations in the first frame, $\mathcal{M}_1 \in \mathbb{R}^{N \times 3}$, we first subtract the median position of the marker observations to center the marker set and remove the global translation offset. Since the markers are unordered, we employ a Point Transformer backbone~\cite{wu2024point} that operates directly on the input point set, without relying on marker identities or layout information. The centered markers are then encoded into point-wise features $\mathbf{F}_1 \in \mathbb{R}^{N \times D}$. Based on these features, we use a transformer decoder with learnable anchor query tokens to predict the positions of a small set of stable anatomical anchors as shown in ~\cref{fig:body_centric}. middle,
\begin{equation}
\mathbf{J}_{1}^{\mathrm{ana}} = \{\mathbf{J}_{1}^{\mathrm{pelvis}}, \mathbf{J}_{1}^{\mathrm{neck}}, \mathbf{J}_{1}^{\mathrm{lhip}}, \mathbf{J}_{1}^{\mathrm{rhip}}, \mathbf{J}_{1}^{\mathrm{lshoulder}}, \mathbf{J}_{1}^{\mathrm{rshoulder}}\} \in \mathbb{R}^{6\times 3},
\label{eq:anatomical_anchors}
\end{equation}
which correspond to semantically meaningful SMPL-H torso joints. We choose these anchors because they are relatively stable under articulation and provide a reliable basis for defining a body-centric canonical space. Specifically, we construct a canonical transformation $\mathcal{T}_{1}$ from these predicted anatomical anchors, which maps the original markers into the body-centric space:
\begin{equation}
\tilde{\mathcal{M}}_{1} = \mathcal{T}_{1}(\mathcal{M}_{1}),
\end{equation}
where $\mathcal{T}_1$ removes the global rotation and translation of the marker observations, with details provided in the \suppl (\cref{Body Centric Canonical Space}). We then use a second Point Transformer backbone to encode the canonicalized marker observations $\tilde{\mathcal{M}}_{1}$ into point-wise features. Based on these features, a transformer decoder predicts the full anchor set $\tilde{\mathbf{A}}_{1}$ in the canonical space. The predicted anchors are subsequently transformed back to the original coordinate system:
\begin{equation}
\mathbf{A}_{1} = \mathcal{T}^{-1}_{1}(\tilde{\mathbf{A}}_{1}) = \{\mathbf{J}_{1}, \mathbf{S}_{1}\} \in \mathbb{R}^{K \times 3},
\end{equation}
where $\mathbf{J}_{1} \in \mathbb{R}^{52 \times 3}$ denotes the SMPL-H body and hand joints, and $\mathbf{S}_{1} \in \mathbb{R}^{61 \times 3}$ denotes surface anchors attached to fixed mesh vertices, giving a total of $K=113$ points. ~\cref{fig:body_centric} visualizes the full anchor set.

We include surface anchors for two reasons: skeletal joints alone cannot fully constrain rotations, as bone-axis rotations induce little or no change in joint positions and thus cause twist ambiguity; off-axis anchors around the limbs and torso resolve this ambiguity during model fitting. Moreover, recovering the SMPL-H shape parameter $\beta$ requires surface, rather than only skeletal, constraints. The selected anchors sparsely but spatially sample body geometry, enabling the solver to recover body proportions, limb thickness, and torso shape.

\subsection{Anchor Trajectory Prediction}
\label{sub:anchor_tracking}
After initialization, we obtain a complete anchor set in the first frame. The goal of this stage is to predict a temporally coherent anchor trajectory over the full sequence. Given marker observations $\mathcal{M}_{1:T}\in\mathbb{R}^{T\times{N}\times{3}}$ and the initialized anchors $\mathbf{A}_1$, we propagate the anchors through time and refine them against the marker observations, yielding the anchor trajectory $\mathbf{A}_{1:T}\in\mathbb{R}^{T\times{K}\times{3}}$.

\begin{figure} \centering
    \includegraphics[width=0.38\textwidth]{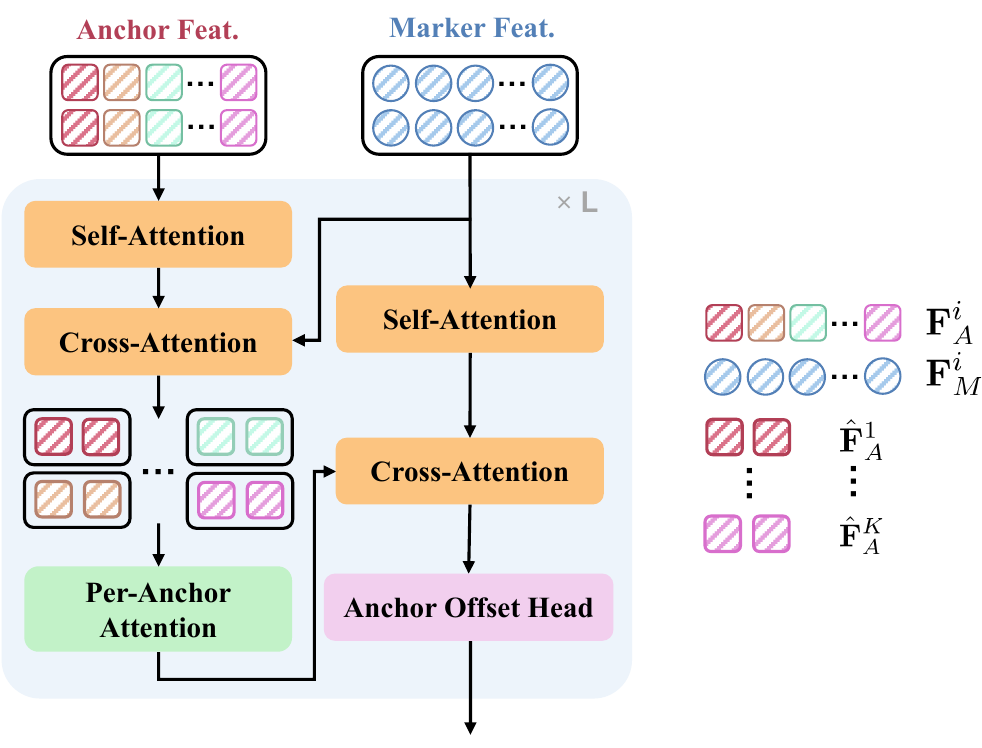}
    \caption{\textbf{Temporal Refinement Network $\Phi$.} The frame-wise attention includes the \textcolor[HTML]{e2b178}{self- and cross- attention}, while the anchor-wise temporal attention contains \textcolor[HTML]{b0d8b7}{per-anchor attention}. This transformer block repeats L times.}
    \label{fig:attention}
\end{figure}

Inspired by recent long-range point tracking methods~\cite{karaev2024cotracker, xiao2024spatialtracker}, we adopt a recurrent sliding-window design with window length $W$ to predict the anchor trajectory over long sequences. 
The initial anchor sequence for the first window, $\mathbf{A}^{\text{init}}_{1:W}$, is obtained by copying the first-frame anchor set $\mathbf{A}_{1}$. 
\textcolor{black}{For subsequent windows, the initial anchors $\mathbf{A}^{\text{init}}_{t:t+W-1}$ inherit the refined anchors from the overlapping frames of the previous window, while the remaining frames repeat the anchor set at the last overlapping frame. The inherited anchors serve only as initialization and are refined again in the new window.}
Based on the stable anatomical anchors $\mathbf{J}_{t}^{\mathrm{ana}}$, which can be obtained either from the anchor initialization module~\cref{sub:anchor_initialization} or by directly selecting the related anchors from $\mathbf{A}_t$ according to the anatomical indices, we construct a body-centric canonical transformation $\mathcal{T}_{t}$ to transform both $\mathbf{A}^{\text{init}}_{t:t+W-1}$ and the marker observations $\mathcal{M}_{t:t+W-1}$:
\begin{equation}
\tilde{\mathbf{A}}^{\text{init}}_{t:t+W-1} = \mathcal{T}_{t}(\mathbf{A}^{\text{init}}_{t:t+W-1}), \quad
\tilde{\mathcal{M}}_{t:t+W-1} = \mathcal{T}_{t}(\mathcal{M}_{t:t+W-1}).
\end{equation}
Using these canonicalized inputs, we concatenate marker observations $\tilde{\mathcal{M}}_{t:t+W-1}$ and initial anchors $\tilde{\mathbf{A}}^{\text{init}}_{t:t+W-1}$ along the point dimension to form a unified set of size $[W, N+K, 3]$. A Point Transformer backbone then extracts point-wise features $\mathbf{F}\in\mathbb{R}^{W\times{(N+K)}\times{D}}$. Joint encoding captures marker-anchor spatial relations, allowing anchors to be refined using nearby marker evidence. We then use the temporal refinement network $\Phi$ in \cref{fig:attention} to predict anchor offsets in canonical space:
\begin{equation}
\Delta \tilde{\mathbf{A}}_{t:t+W-1} = \Phi(\mathbf{F}_{M}, \mathbf{F}_{A}),
\end{equation}
where the marker features $\mathbf{F}_{M} \in \mathbb{R}^{W \times N \times D}$ and the anchor features $\mathbf{F}_{A} \in \mathbb{R}^{W \times K \times D}$ are obtained by splitting the point-wise features $\mathbf{F}$.
The refined anchor trajectory in canonical space is obtained as
\begin{equation}
\tilde{\mathbf{A}}_{t:t+W-1} = \tilde{\mathbf{A}}^{\text{init}}_{t:t+W-1} + \Delta \tilde{\mathbf{A}}_{t:t+W-1}.
\end{equation}
We then transform the refined anchors back to the original coordinate system using the inverse canonical transformation:
\begin{equation}
\mathbf{A}_{t:t+W-1} = \mathcal{T}^{-1}_{t}\!\left(\tilde{\mathbf{A}}_{t:t+W-1}\right),
\end{equation}
where $\mathcal{T}_{t}$ denotes the canonical transformation defined by the first anchor state of the window. The resulting anchors in the original space are then used to construct the initialization for the subsequent window. This allows tracking information to propagate through arbitrarily long sequences while keeping memory bounded.

Within each window, the temporal refinement network $\Phi$ alternates between frame-wise attention and anchor-wise temporal attention. In frame-wise attention, self- and cross-attention operate on the anchor features $\mathbf{F}_{A}^{i} \in \mathbb{R}^{K \times D}$ and marker features $\mathbf{F}_{M}^{i} \in \mathbb{R}^{N \times D}$ of each frame $i$, allowing anchors to gather evidence from markers in the same frame. In anchor-wise temporal attention, we reshape $\mathbf{F}_{A}$ into $\hat{\mathbf{F}}_{A} \in \mathbb{R}^{K \times W \times D}$ and apply self-attention along time to each anchor trajectory $\hat{\mathbf{F}}_{A}^{j} \in \mathbb{R}^{W \times D}$. This lets information flow across frames through ordered anchor tokens, while avoiding the unstable temporal alignment of unordered markers. The resulting anchor features are passed to an MLP head to regress the anchor offsets $\Delta \tilde{\mathbf{A}}_{t:t+W-1}$. We also maintain a hidden state across consecutive windows, which refines the point-wise features of subsequent windows through self- and cross-attention.

\subsection{Differentiable and Learnable Solver}
\label{sub:differentiable_solver}

Given the tracked anchor trajectory $\mathbf{A}_{1:T}$, the final stage of our method solves for the SMPL-H parameters in sliding windows, following the same windowed formulation used for anchor trajectory prediction. For each window, we optimize the pose $\theta_{t:t+W-1}\in\mathbb{R}^{W\times156}$, global translation $\tau_{t:t+W-1}\in\mathbb{R}^{W\times3}$, and a sequence-level shape parameter $\beta\in\mathbb{R}^{10}$. Here, the 156-dimensional pose vector represents the axis-angle rotations of 52 SMPL-H joints. These parameters are initialized to zeros for each window, i.e., $\theta_{t:t+W-1} = 0$, $\tau_{t:t+W-1} = 0$, and $\beta = 0$. Since the anchors maintain fixed correspondences to the SMPL-H model, they provide direct geometric constraints for parametric body fitting. We formulate this stage as a nonlinear least-squares problem and optimize the SMPL-H parameters using a differentiable Gauss--Newton solver. 
Specifically, for each window, we define the nonlinear least-squares objective as
\begin{equation}
\begin{split}
E(\theta_{t:t+W-1}, \tau_{t:t+W-1}, \beta)
= & \quad E_{\mathrm{obs}}(\theta_{t:t+W-1},\tau_{t:t+W-1}, \beta) \\ 
+ & \quad E_{\mathrm{prior}}(\theta_{t:t+W-1}) \\
+ & \quad E_{\mathrm{smooth}}(\theta_{t:t+W-1},\tau_{t:t+W-1}, \beta),
\label{eq:total_objectives}
\end{split}
\end{equation}
where the observation term $E_{\mathrm{obs}}$ aligns the SMPL-H joints and surface anchors with the tracked anchor trajectory within the window,
\begin{equation}
    E_{\mathrm{obs}} = \sum_{i=t}^{t+W-1}\|w^{c}_{i}*(\hat{\mathbf{A}}(\theta_{i}, \tau_{i}, \beta) -\mathbf{A}_{i}) \|^{2}.
\label{eq:obs_objects}
\end{equation}
Here, $\hat{\mathbf{A}}(\theta_{i}, \tau_{i}, \beta)$ represents the SMPL-H joints and surface anchors at frame $i$, computed by applying the SMPL-H model with the pose $\theta_{i}$, translation $\tau_{i}$, and shape $\beta$. The model deformation is based on pose- and shape-dependent blend shapes and Linear Blend Skinning (LBS), with further details provided in the \suppl (\cref{Parametric Model: SMPL-H}). The resulting joint locations and surface anchor positions are used to minimize the discrepancy with the tracked anchor trajectory. The weight term \( w^{c}\in\mathbb{R}^{W\times{K}} \) represents a learnable confidence that adjusts the contribution of each observation error based on the reliability of the corresponding anchor.

To ensure temporal smoothness of the anchor trajectory $\hat{\mathbf{A}}_{t:t+W-1}$, we introduce a smoothness term $E_{\mathrm{smooth}}$, which penalizes large changes in the velocity of the anchors between consecutive frames. The smoothness term is defined as:
\begin{equation}
\begin{gathered}
\Delta^{2}\hat{\mathbf{A}}_{i}
\;=\; \hat{\mathbf{A}}(\theta_{i-1}, \tau_{i-1}, \beta)
- 2\hat{\mathbf{A}}(\theta_{i}, \tau_{i}, \beta)
\; + \hat{\mathbf{A}}(\theta_{i+1}, \tau_{i+1}, \beta), \\
E_{\mathrm{smooth}}
\;=\; \sum_{i=t+1}^{t+W-2}
\left\|w^{s}_{i} * \Delta^{2}\hat{\mathbf{A}}_{i}\right\|^{2}.
\end{gathered}
\label{eq:smooth_objects}
\end{equation}
where \( w^{s}\in\mathbb{R}^{W\times{K}} \) is a learnable smoothness weight that modulates the influence of the smoothness term.

The prior term $E_{\mathrm{prior}}$ penalizes deviations of the pose from the prior distribution, encouraging realistic body configurations.
\begin{equation}
    E_{\mathrm{prior}} = \sum_{i=t}^{t+W-1}-\log(w^{p}_{i} * P(\theta_{i})),
    \label{eq:prior_obj}
\end{equation}
where \(P(\theta_{i})\) is the probability that pose \(\theta_i\) follows the prior, and $w^{p} \in \mathbb{R}^{W \times 156}$ is a learnable per-frame weight. This term regularizes the optimization toward realistic human poses.

We optimize the above objective with an iterative Gauss--Newton procedure. We first describe the update induced by the observation term ~\cref{eq:obs_objects}. For each window, we collect the unknown SMPL-H parameters into a single vector,
\begin{equation}
\mathbf{x} = [\theta_{t:t+W-1}, \tau_{t:t+W-1}, \beta] \in \mathbb{R}^{W*(156+3)+10}.
\end{equation}
The observation term in ~\cref{eq:obs_objects} can be written as a weighted nonlinear least-squares problem:
\begin{equation}
E_{\mathrm{obs}}(\mathbf{x}) = \sum_{i=t}^{t+W-1}
\|
\mathbf{r}^{c}_{i}(\mathbf{x})
\|^{2}, \text{where} ~
\mathbf{r}^{c}_{i}(\mathbf{x})
=
w^{c}_{i} *
(
\hat{\mathbf{A}}(\theta_i,\tau_i,\beta) - \mathbf{A}_{i}
).
\end{equation}
At each Gauss--Newton iteration $it$, we linearize the residual around the current estimate $\mathbf{x}^{(it)}$:
\begin{equation}
\mathbf{r}^{c}_{i}(\mathbf{x}^{(it)}+\Delta \mathbf{x})
\approx
\mathbf{r}^{c}_{i}(\mathbf{x}^{(it)})
+\mathcal{J}^{c,(it)}_{i}\Delta \mathbf{x},
\end{equation}
where
\begin{equation}
\mathcal{J}^{c,(it)}_{i}
=
\left.
\frac{\partial \mathbf{r}^{c}_{i}}{\partial \mathbf{x}}
\right|_{\mathbf{x}^{(it)}}
=
\left.
\frac{\partial \mathbf{r}^{c}_{i}}{\partial \hat{\mathbf{A}}_{i}}
\left[
\frac{\partial \hat{\mathbf{A}}_{i}}{\partial \theta_{i}},
\frac{\partial \hat{\mathbf{A}}_{i}}{\partial \tau_{i}},
\frac{\partial \hat{\mathbf{A}}_{i}}{\partial \beta}
\right]
\right|_{\mathbf{x}^{(it)}} .
\end{equation}
Here, $\mathcal{J}^{c,(it)}_{i}$ is the Jacobian of the weighted anchor residual with respect to the optimized SMPL-H parameters. Its computation follows the SMPL-H kinematic chain and the derivatives of the Linear Blend Skinning function; we provide the detailed derivation in the \suppl (\cref{Jacobian of SMPL-H Anchors}).
Stacking the residuals and Jacobians over all frames in the window gives the following linear system:
\begin{equation}
\left(\sum_{i=t}^{t+W-1}(\mathcal{J}_{i}^{c,(it)})^{\top}\mathcal{J}_{i}^{c,(it)}\right)\Delta \mathbf{x}
= -\sum_{i=t}^{t+W-1}(\mathcal{J}_{i}^{c,(it)})^{\top}\mathbf{r}_{i}^{c}(\mathbf{x}^{(it)}).
\label{eq:obs_linear_sys}
\end{equation}

The smoothness term in ~\cref{eq:smooth_objects} is treated analogously. We incorporate the smoothness residual $\mathbf{r}^{s}_{i}(\mathbf{x})$, with the corresponding Jacobian 
$\mathcal{J}_{i}^{s,(it)}=\left.\frac{\partial \mathbf{r}^{s}_{i}}{\partial \mathbf{x}}\right|_{\mathbf{x}^{(it)}}$, into this linear system. The prior term in~\cref{eq:prior_obj} is handled differently. We use DPoser-X~\cite{lu2025dposer} as the pose prior, which models whole-body human poses with a diffusion model. Unlike the observation and smoothness terms, this prior does not provide an explicit least-squares residual with an analytical Jacobian. Instead, the diffusion model provides a denoising direction for the current pose estimate, which can be interpreted as a score-based direction toward the pose prior distribution. We therefore incorporate this direction directly into the right-hand side of the Gauss--Newton linear system:
\begin{equation}
\begin{split}
\left(
\sum_{\mathrm{y}\in\{c,s\}}
(\mathcal{J}^{\mathrm{y},(it)})^{\top}
\mathcal{J}^{\mathrm{y},(it)}
\right)
& \Delta \mathbf{x}
= \\
-
\sum_{\mathrm{y}\in\{c,s\}}
(\mathcal{J}^{\mathrm{y},(it)})^{\top} &
\mathbf{r}^{\mathrm{y}}(\mathbf{x}^{(it)})
+
w^{p}*\boldsymbol{\epsilon}_{prior}(\mathbf{x}^{(it)})
\label{eq:final_linear_sys}
\end{split}
\end{equation}
where $\mathrm{y}\in\{c,s\}$ denotes the observation and smoothness terms, $\boldsymbol{\epsilon}_{\mathrm{prior}}(\mathbf{x}^{(it)})$ denotes the denoising direction at iteration $it$. %

We solve this linear system efficiently in ~\cref{eq:final_linear_sys} using a parallel conjugate gradient (PCG) method. The resulting update $\Delta \mathbf{x}$ is then applied to the current estimate:
\begin{equation}
\mathbf{x}^{(it+1)} = \mathbf{x}^{(it)} + \Delta \mathbf{x}.
\end{equation}

Through the above formulation, the entire iterative Gauss--Newton procedure is differentiable. We introduce a weight prediction network $\Psi$ to estimate the learnable weights:
\begin{equation}
(w_{\Psi}^{c}, w_{\Psi}^{s}, w_{\Psi}^{p}) = \Psi(\mathbf{A}_{t:t+W-1}),
\end{equation}
where $w_{\Psi}^{c} \in \mathbb{R}^{N_{\mathrm{it}}\times W\times K}$ contains the observation confidence for all Gauss--Newton iterations, and $N_{\mathrm{it}}$ denotes the number of solver iterations. At iteration $it$, the observation confidence used in the linear system is the corresponding slice $w^{c}=w_{\Psi}^{c,(it)}\in\mathbb{R}^{W\times K}$. The same indexing is used for the smoothness and prior weights $w_{\Psi}^{s}$ and $w_{\Psi}^{p}$. The network $\Psi$ uses a Point Transformer backbone followed by transformer decoders to infer the observation confidence, smoothness weight, and prior weight from the tracked anchor trajectory. It follows a similar architecture to the temporal refinement network $\Phi$, but uses separate prediction heads to regress the learnable weights instead of anchor offsets. Since the linear system, PCG method, and parameter update are all differentiable, supervision on $\mathbf{x}^{(it)}$ can be back-propagated through each Gauss--Newton iteration to train $\Psi$. The update rule can therefore be written as
\begin{equation}
\mathbf{x}^{(it+1)} = \mathbf{x}^{(it)} + \Delta \mathbf{x}(w_{\Psi}^c, w_{\Psi}^s, w_{\Psi}^p; \mathbf{A}_{t:t+W-1}).
\end{equation}
This end-to-end formulation allows the solver to learn input-dependent weighting strategies for anchor correspondences, temporal smoothness, and pose priors. Consequently, the observation confidence, smoothness strength, and prior strength can be dynamically adjusted for each input sequence as shown in \cref{fig:learnable_weight}, enabling the optimization process to be guided by the final SMPL-H reconstruction objective rather than by fixed hand-designed weights.

\subsection{Supervision}
We train the proposed framework with supervision at both the anchor and parameter levels, corresponding to the three stages of our pipeline. Let $\mathbf{A}^{*}_{1:T}$ represent the ground-truth anchor trajectory derived from the ground-truth SMPL-H parameters, and $(\theta^{*}_{1:T}, \tau^{*}_{1:T}, \beta^{*})$ denote the ground-truth pose, translation, and shape.

\noindent\textbf{Anchor Initialization.}
In the initialization stage, we supervise the stable anatomical anchors by minimizing the L1 loss between the predicted anatomical anchors $\mathbf{J}_{1}^{\mathrm{ana}}$ and the ground-truth anchors $\mathbf{J}_{1}^{\mathrm{ana*}}$:
$\mathbf{A}^{*}_{1}$: $\mathcal{L}_{\mathrm{ana}} =
\left\|
\mathbf{J}^{\mathrm{ana}}_{1} - \mathbf{J}^{\mathrm{ana}*}_{1}
\right\|_{1}$.
We also supervise the predicted full anchor set $\mathbf{A}_{1}$ by minimizing the L1 loss with the ground-truth anchor set $\mathbf{A}^{*}_{1}$: 
$ \mathcal{L}_{\mathrm{init}} = \left\|
\mathbf{A}_{1} - \mathbf{A}^{*}_{1}
\right\|_{1}.$

\noindent\textbf{Anchor Trajectory Prediction.}
For temporal anchor refinement, we supervise the predicted anchor trajectory over each window by minimizing the L1 loss between the predicted and ground-truth anchor positions: $\mathcal{L}_{\mathrm{track}} =
\sum_{i=t}^{t+W-1}
\left\|
\mathbf{A}_{i} - \mathbf{A}^{*}_{i}
\right\|_{1}.$

\noindent\textbf{Differentiable and Learnable Solver.}
In the final stage, we supervise the SMPL-H anchors obtained at each Gauss--Newton iteration $it$. Given the current parameters 
$\theta^{(it)}_i$, $\tau^{(it)}_i$, and $\beta^{(it)}$, we compute the corresponding SMPL-H anchors 
$\hat{\mathbf{A}}(\theta^{(it)}_i, \tau^{(it)}_i, \beta^{(it)})$ and apply anchor, smoothness, and prior losses:
\begin{equation}
\begin{split}
\mathcal{L}_{\mathrm{anchor}} &= 
\sum_{it}\sum_{i=t}^{t+W-1}
\|\hat{\mathbf{A}}(\theta^{(it)}_i, \tau^{(it)}_i, \beta^{(it)}) - \mathbf{A}^{*}_i\|_1, \\
\mathcal{L}_{\mathrm{smooth}} &= \|\Delta^{2}\hat{\mathbf{A}}_{i}^{(it)} \|_1, \quad
\mathcal{L}_{\mathrm{prior}} = \sum_{it}\sum_{i=t}^{t+W-1} - \log{(P(\theta^{(it)}_{i}))}
\end{split}
\end{equation}

\subsection{Implementation Details}
\noindent\textbf{Architecture.}
We process sequences using recurrent windows of length $W=16$ and maintain a 768-dimensional hidden state across windows. Consecutive windows overlap by $W/2$ frames. The Point Transformer backbones use encoder channels $[32,64,128,256,512]$ and decoder channels $[256,256,256,256]$. We unroll the differentiable solver for 10 iterations and compute the SMPL-H kinematic-chain Jacobians using our CUDA implementation.

\noindent\textbf{Anchor Selection.}
We use the 86-point Superset marker configuration as a candidate pool and select 51 surface locations to provide spatial coverage across the body. The selection generally retains at least two locations per body region, with four on each foot and two on each lower leg and shoulder region, to constrain body shape and local twisting. We further add one surface vertex at each of the ten fingertips, yielding 61 surface anchors. Together with the 52 SMPL-H joints, they form our default set of 113 anchors.

\noindent\textbf{Marker Configurations and Augmentation.}
Each training batch uses one real or randomly generated marker configuration, with 38--175 markers across configurations. We collect 75 real configurations used in practical MoCap settings from the SOMA project~\cite{ghorbani2021soma}, such as Mixamo and CMU-II; those without hand markers are supplemented with sampled SMPL-H hand vertices. Random configurations sample SMPL-H vertices by body region (e.g., two to three markers per finger), and the selected vertex IDs define the observations. We augment them with marker shuffling, spatial jitter, fixed marker-placement offsets, temporal drift, marker dropout, outlier perturbations, and ghost-marker insertion under their respective sampling protocols. For example, outlier noise is applied with probability 0.5 and affects 30\% of the markers in 25\% of the frames.

\noindent\textbf{Training Strategy.}
We train the three pipeline stages progressively. We first train the anchor initialization module with first-frame supervision, then add the trajectory prediction module and train it on 24-frame clips. After convergence, we increase the clip length to 56 before adding the differentiable solver and training the full pipeline end-to-end. 

\section{Experiments}

\subsection{Evaluation Protocols}

\noindent\textbf{Synthetic Evaluation.}
Following LocalMoCap~\cite{pan2023locality}, we generate unordered,
noisy markers by sampling SMPL-H vertices under different marker
configurations and noise levels, using CMU MoCap body
motions~\cite{cmu_mocap} and GRAB hand motions~\cite{taheri2020grab}.
The dataset contains 2,068 sequences (mean/max length:
1,687/22,948 frames), with the first 80\% used for training and
the remaining 20\% for testing.
We compare against SOMA~\cite{ghorbani2021soma},
LocalMoCap~\cite{pan2023locality}, and
OpenMoCap~\cite{qian2025openmocap}, all trained on CMU MoCap.
SOMA's marker-label classification and the fixed network input
sizes of LocalMoCap and OpenMoCap make these baselines
configuration-specific; each is therefore evaluated on test data
generated using its corresponding marker layout.
Our method uses a single trained model across all layouts.
We report mean per joint position error (MPJPE), the mean
Euclidean error over body and hand joints, and mean per vertex
position error (MPVPE), the corresponding error over SMPL-H
surface vertices, to assess pose and surface reconstruction.

\noindent\textbf{Real-World Evaluation.}
Following AMASS~\cite{AMASS:ICCV:2019}, we evaluate on the
Synchronized Scans and Markers (SSM) dataset, which pairs optical
marker trajectories with temporally aligned 4D body scans.
We compare against MoSh++~\cite{AMASS:ICCV:2019}, the
model-fitting method used in AMASS, using the same 67-marker
coordinates: MoSh++ receives known marker-to-body correspondences,
while our method discards semantic labels and processes the
coordinates as unordered point sets without predefined
correspondences.
Since ground-truth SMPL-H parameters are unavailable, the scans
serve solely as reference surfaces for quantitative evaluation.
We report symmetric Chamfer distance between points sampled from
the reconstructed mesh and the synchronized scan, and valid
marker-to-mesh distance, the mean closest distance from valid
markers to the reconstructed mesh surface.
Valid markers lie within 1 cm of the synchronized scan surface
and account for 97\% of all markers.

\subsection{Experiment Results}

\begin{figure*}[thbp] \centering
    \includegraphics[width=0.93\textwidth]{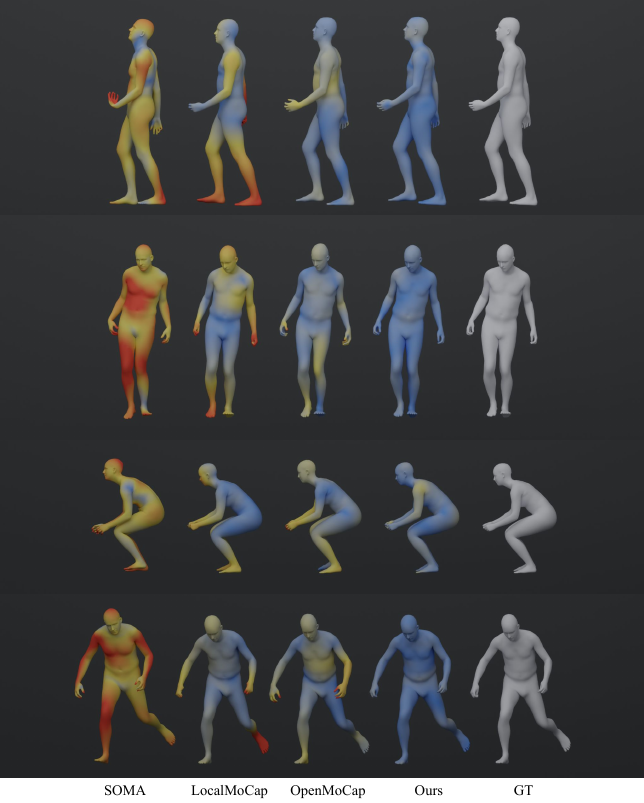}
    \caption{\textbf{Qualitative results.} We use vertex-to-vertex error to visualise the deformed SMPL-H model based on the estimated pose, global translation, and shape parameters. Since LocalMocap~\cite{pan2023locality} and OpenMoCap~\cite{qian2025openmocap} do not predict shape, we use the ground-truth shape values for these methods. Additionally, SOMA~\cite{ghorbani2021soma} and OpenMoCap do not predict hand pose, for which we adopt the mean hand pose. }
    \label{fig:qualitative}
\end{figure*}

\noindent\textbf{Comparison on Synthetic Data.}
We evaluate our method under two noise settings and multiple marker configurations. The first follows LocalMoCap~\cite{pan2023locality} with occlusion noise, and the second follows SOMA~\cite{ghorbani2021soma} with both occlusion and outlier noise. In each setting, we generate test sequences using the marker configurations of SOMA~\cite{ghorbani2021soma}, LocalMoCap~\cite{pan2023locality}, and OpenMoCap~\cite{qian2025openmocap}. Because LocalMoCap and OpenMoCap require ordered marker inputs, their results are shown in gray in \cref{tab:table1}; SOMA and our method, by contrast, operate on unordered markers.
As shown in \cref{tab:table1}, a single model generalizes across all marker configurations, yielding lower MPJPE than SOMA and matching or outperforming LocalMoCap and OpenMoCap on their respective configurations. The lower MPVPE relative to SOMA further confirms that the anchor representation effectively constrains both pose and surface recovery. Additional qualitative results are provided in \cref{fig:qualitative}.

\begin{table}[t]
  \centering
  \caption{\textbf{Quantitative comparison across noise settings and marker configurations.} The two groups represent data w/ occlusion only and w/ both occlusion and outliers. Configurations 1--3 correspond to SOMA, LocalMoCap, and OpenMoCap. \textcolor{gray}{Gray methods} require ordered marker inputs. MPJPE is reported as body/hand error when available; all errors are in millimetres. For ours, joints and vertices are computed by deforming SMPL-H using the final optimized pose, global translation, and shape parameters.}
  \label{tab:table1}
  \resizebox{\columnwidth}{!}{
    \begin{tabular}{c|c|*{3}{c}|*{3}{c}}
      \bottomrule
      \multirow{2}{*}{Metric} & \multirow{2}{*}{Method}
      & \multicolumn{3}{c|}{Data w/ occlusion}
      & \multicolumn{3}{c}{Data w/ occlusion and outlier} \\
      &
      & config. 1 & config. 2 & config. 3
      & config. 1 & config. 2 & config. 3 \\
      \midrule
      \multirow{4}{*}{MPJPE(mm)}
      & SOMA
      & 13.73/- & -/- & -/-
      & 16.63/- & -/- & -/- \\
      & \textcolor{gray}{LocalMocap} %
      & -/- & \textcolor{gray}{\textbf{7.80}}/\textcolor{gray}{16.08} & -/-
      & -/- & \textcolor{gray}{11.30}/\textcolor{gray}{28.12} & -/- \\
      & \textcolor{gray}{OpenMoCap} %
      & -/- & -/- & \textcolor{gray}{16.00}/-
      & -/- & -/- & \textcolor{gray}{16.70}/- \\
      & Ours
      & \textbf{7.96}/- & 8.03/\textbf{7.47} & \textbf{9.01}/-
      & \textbf{8.48}/- & \textbf{8.55}/\textbf{9.07} & \textbf{10.06}/- \\
      \hline
      \multirow{2}{*}{MPVPE(mm)}
      & SOMA
      & 20.81 & -- & --
      & 24.85 & -- & -- \\
      & Ours
      & \textbf{9.41} & \textbf{9.34} & \textbf{10.35}
      & \textbf{10.16} & \textbf{10.20} & \textbf{11.92} \\
      \toprule
    \end{tabular}
  }
\end{table}

\noindent\textbf{Comparison on Real-World Data.}
\begin{table}[t]
  \centering
  
  \caption{\textbf{Comparison on the real-world SSM dataset.} Marker-specific prior denotes access to marker-to-body correspondences and a marker-specific surface-offset constraint. MoSh++ is provided with both, whereas our method receives only the same 67-marker coordinates and uses neither. All distances are reported in millimeters; lower is better.}
  \label{tab:table_ssm}
  \resizebox{0.47\textwidth}{!}{
    \begin{tabular}{lccc}
      \toprule
      Method & Marker-specific Prior & Symmetric Chamfer & Valid Marker-to-Mesh \\
      \midrule
      MoSh++ & \cmark & \textbf{11.3} & 10.0 \\
      Ours & \xmark & 15.3 & \textbf{7.5} \\
      \bottomrule
    \end{tabular}
  }
\end{table}

\begin{figure}[t] \centering
  \includegraphics[width=0.48\textwidth]{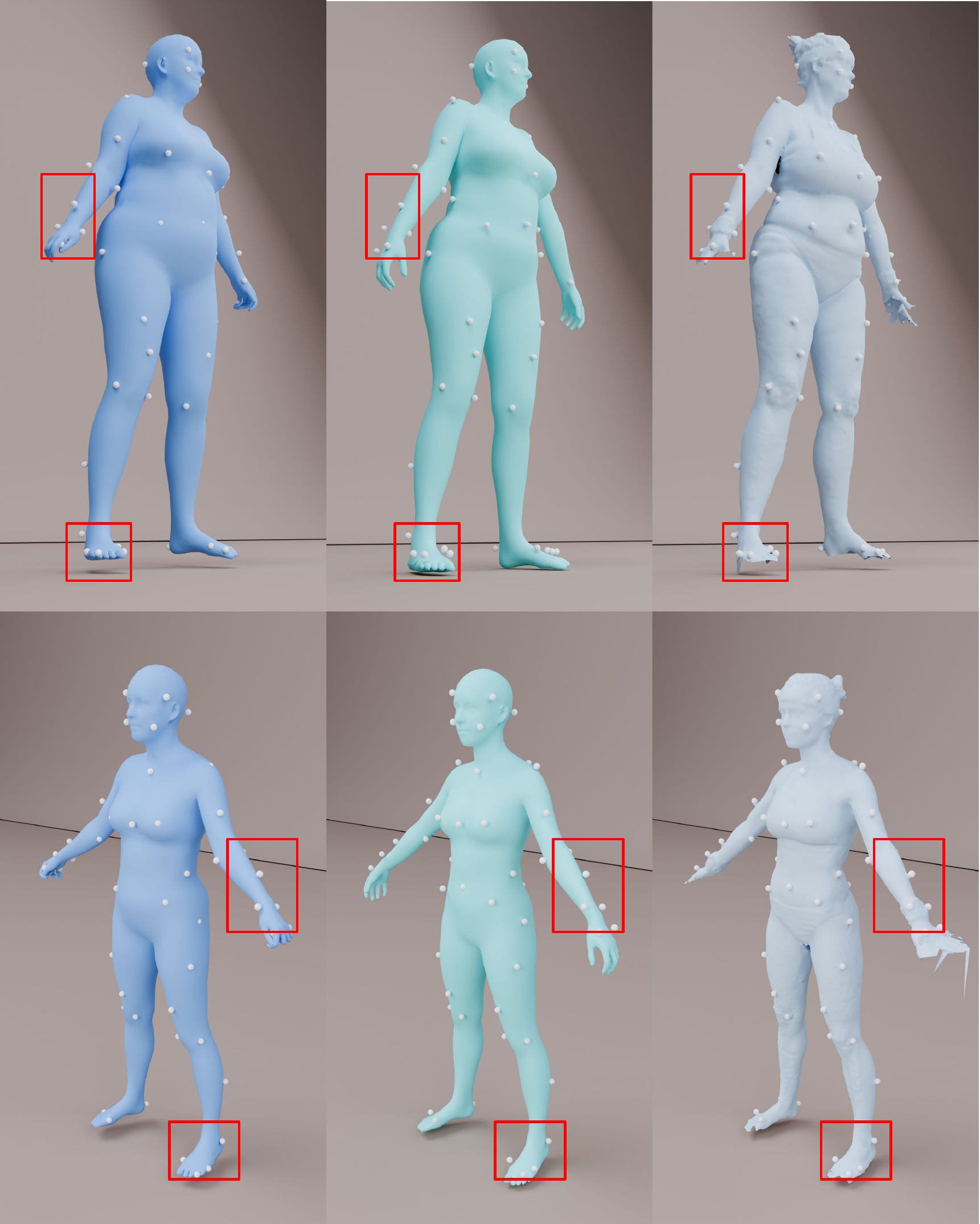}
  \caption{\textbf{Comparison on SSM.} From left to right: \textcolor[HTML]{6F95C5}{our reconstruction}, the \textcolor[HTML]{79C7C9}{MoSh++ reconstruction}, and the \textcolor{gray}{synchronized scan}, all overlaid with the same 67 marker observations (white spheres). Unlike MoSh++, our method uses neither marker-to-body correspondences nor marker-specific surface-offset constraints.}
  \label{fig:ssm_results}
\end{figure}

As shown in \cref{tab:table_ssm}, MoSh++ achieves a lower symmetric Chamfer distance than our method (11.3 mm versus 15.3 mm). MoSh++ is provided with marker-to-body correspondences and explicitly regularizes its latent markers to maintain a prescribed mean offset of 9.5 mm from the body surface~\cite{AMASS:ICCV:2019}. In contrast, our method receives the same 67-marker coordinates without semantic labels, predefined correspondences, or a marker-specific surface-offset constraint. Although its symmetric Chamfer distance is 4.0 mm higher, our method reduces the valid marker-to-mesh distance from 10.0 mm to 7.5 mm, indicating that the reconstructed surface lies closer to the valid marker observations. These complementary results, illustrated in \cref{fig:ssm_results}, show that MoSh++ more closely matches the scan geometry, whereas our configuration-agnostic formulation follows the raw marker evidence more closely without relying on marker-specific prior information.
Since the SSM layout contains no finger markers, body motion remains recoverable, whereas the hands fall back to plausible average poses guided by the pose prior and temporal smoothness rather than recovering fine-grained finger articulation.

\begin{table}[t]
\centering
\caption{\textbf{Pipeline component analysis.} Stage-wise MPJPE is evaluated on the same 52 SMPL-H joints under \texttt{config.2} with occlusion and outlier noise. Runtime values are per-frame averages.}
\label{tab:runtime_analysis_transposed}
\resizebox{0.44\textwidth}{!}{
\begin{tabular}{lccc}
\toprule
\textbf{Metric} & \textbf{Anchor Init} & \textbf{Trajectory Prediction} & \textbf{Solver} \\
\midrule
MPJPE (mm) & 16.2 & 7.6 & 8.8 \\
Runtime (ms/frame) & 3.08 & 12.22 & 23.03 \\
\bottomrule
\end{tabular}
}
\end{table}

\begin{table}[t]
\centering
\caption{\textbf{Ablation of learnable solver weights.} Sparse outliers shift 10\% of anchor observations by 0.86--1.68 m, while regional outliers corrupt a randomly selected body part in 25\% of the frames. Vanilla LM uses uniform observation weights. All variants run for 10 iterations, and MPJPE is evaluated on the same 52 SMPL-H joints.}
\label{tab:learnable_solver_weights}
\resizebox{0.44\textwidth}{!}{
\begin{tabular}{llccc|c}
\toprule
\multirow{2}{*}{Outliers} & \multirow{2}{*}{Solver} & \multicolumn{3}{c|}{Learnable Weights} & \multirow{2}{*}{MPJPE (mm) $\downarrow$} \\
& & $w^c$ & $w^s$ & $w^p$ & \\
\midrule
\multirow{2}{*}{Sparse}
& Vanilla LM & \xmark & \xmark & \xmark & 31.0 \\
& Ours       & \cmark & \xmark & \xmark & \textbf{1.5} \\
\midrule
\multirow{4}{*}{Sparse + Regional}
& Ours & \cmark & \xmark & \xmark & 11.5 \\
& Ours & \cmark & \cmark & \xmark & 2.1 \\
& Ours & \cmark & \xmark & \cmark & 6.4 \\
& Ours & \cmark & \cmark & \cmark & \textbf{1.9} \\
\bottomrule
\end{tabular}
}
\end{table}

\subsection{Ablation Studies and Analysis}
\noindent\textbf{Pipeline Component Analysis.}
As shown in \cref{tab:runtime_analysis_transposed}, anchor initialization obtains an MPJPE of 16.2 mm, which trajectory prediction reduces to 7.6 mm. The solver yields 8.8 mm while converting the anchors into temporally smooth SMPL-H motion. The three stages take 3.08, 12.22, and 23.03 ms per frame, respectively, totaling 38.33 ms per frame.

\begin{figure*}[!t]
    \centering
    \includegraphics[width=0.98\textwidth]{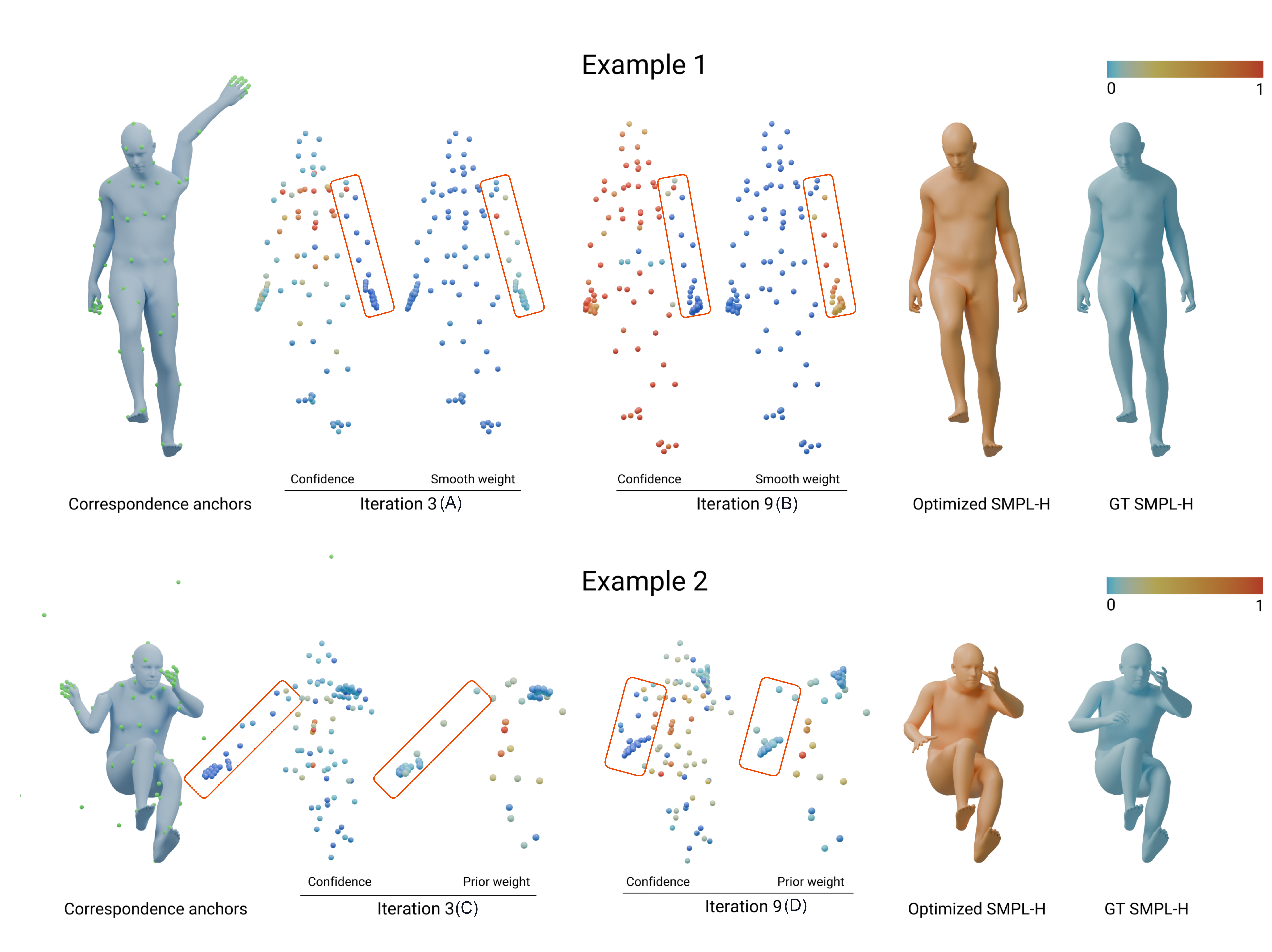}
    \caption{\textbf{Effect of learnable weights in the differentiable solver.}
The noise-corrupted \textcolor[HTML]{7EC971}{correspondence anchors} are provided as input to the solver, which estimates the final SMPL-H parameters from the T-pose. The left-hand correspondences are distorted in Example~1, whereas the right-hand correspondences are distorted in Example~2. Both examples learn observation confidence, together with smoothness weights in Example~1 and pose-prior weights in Example~2. Panels (A)--(D) visualize intermediate SMPL-H estimates at the third and ninth Gauss--Newton iterations, sampled at the anchor vertex IDs; \textcolor[HTML]{559EB8}{cool}-to-\textcolor[HTML]{AF3C27}{warm} colors encode increasing normalized values of the corresponding weights. (A) At iteration~3 in Example~1, the incorrect left-hand anchors receive low confidence but high smoothness weights, allowing valid correspondences in neighboring frames to guide recovery of the hand pose. (B) This weighting pattern persists at iteration~9, while confidence on the other, correctly aligned body parts increases as optimization proceeds. (C) At iteration~3 in Example~2, the incorrect right-hand anchors receive low confidence, while their prior weights are slightly higher than the confidence weights. The resulting prior-dominated updates pull the right hand toward the mean pose encoded by the learned prior, thereby preventing distortion. (D) This weighting pattern persists at iteration~9, while confidence on the other, correctly aligned body parts increases as optimization proceeds.}
    \label{fig:learnable_weight}
\end{figure*}

\noindent\textbf{Differentiable and Learnable Solver.}
Using ground-truth anchor trajectories, we compare all variants over 10 iterations with MPJPE evaluated on the same 52 SMPL-H joints. As shown in \cref{tab:learnable_solver_weights}, learned confidence reduces the error under sparse outliers from 31.0 mm with vanilla LM to 1.5 mm. Under combined sparse and regional outliers, adding smoothness and the pose prior improves MPJPE from 11.5 mm to 1.9 mm, demonstrating their complementary robustness.
To assess optimization efficiency, we compare our solver with the Levenberg--Marquardt (LM) optimizer used in ETCH-X~\cite{li2026etch}; both receive the same predicted anchors and optimize the same SMPL-H parameters. With only 10 iterations, our solver achieves competitive accuracy (\cref{tab:table_theseus_LM}), indicating efficient and robust optimization. \cref{fig:learnable_weight} illustrates the learnable weights.

\begin{table}[t]
\centering
\caption{\textbf{Comparison of our differentiable, learnable solver and a Levenberg--Marquardt (LM) solver~\cite{li2026etch} for SMPL-H fitting.} Iterations denote optimization steps. All reported metrics are in millimeters.
}
\label{tab:table_theseus_LM}
\resizebox{0.44\textwidth}{!}{
\begin{tabular}{c|c|*{4}{c}}
\bottomrule
\multirow{2}{*}{Solver} & \multirow{2}{*}{Iterations} &
\multicolumn{4}{c}{Data w/ occlusion and outlier} \\
       & & MPJPE $\downarrow$ & Body MPJPE $\downarrow$ & Hand MPJPE $\downarrow$ & MPVPE $\downarrow$\\
\hline
\multirow{5}{*}{LM} & 80 & 7.64 & 7.73 & 7.57 & 9.47 \\
                           & 40 & 8.34 & 8.33 & 8.34 & 10.62 \\
                           & 30 & 8.91 & 8.76 & 9.02 & 11.48 \\
                           & 20 & 10.15 & 9.98 & 10.28 & 13.71 \\
                           & 10 & 38.63 & 35.07 & 41.24 & 40.60 \\
\hline
Ours & 10 & 8.76 & 8.48 & 8.97 & 10.12 \\
\toprule
\end{tabular}
}
\end{table}

\noindent\textbf{Anchor and Initialization Design.}
As shown in \cref{tab:anchor_initialization}, 113 anchors provide the best balance between joint and surface accuracy, supporting our default choice of 52 joints and 61 surface anchors. Independently applying the initializer to each frame yields an MPJPE of 22.2 mm, compared with 16.2 mm for our first-frame initialization, while the coarse-to-fine design consistently outperforms one-step prediction. Replacing the coarse-to-fine initializer with one-step prediction further increases trajectory/solver MPJPE from 7.6/8.8 mm reported in \cref{tab:runtime_analysis_transposed} to 9.1/10.6 mm.

\begin{table}[t]
\centering
\caption{\textbf{Anchor and initialization design.} (a) Total anchor counts, including the same 52 SMPL-H joints, are evaluated using the confidence-only solver with ground-truth correspondences under sparse outliers. (b) Initialization MPJPE is evaluated under \texttt{config.2} with occlusion and outlier noise; per-frame initialization independently predicts anchors in every frame. All errors are in millimeters.}
\label{tab:anchor_initialization}
\scriptsize
\begin{minipage}[t]{0.35\columnwidth}
\centering
\textbf{(a) Anchor-set size}\par\smallskip
\setlength{\tabcolsep}{1.5pt}
\begin{tabular}{lcc}
\toprule
Anchors & MPJPE $\downarrow$ & MPVPE $\downarrow$ \\
\midrule
86  & 2.1 & 9.1 \\
99  & 2.0 & 3.8 \\
113 & \textbf{1.5} & \textbf{2.5} \\
148 & 1.8 & \textbf{2.5} \\
\bottomrule
\end{tabular}
\end{minipage}
\hfill
\begin{minipage}[t]{0.58\columnwidth}
\centering
\textbf{(b) Initialization strategy}\par\smallskip
\setlength{\tabcolsep}{2.5pt}
\begin{tabular}{lcc}
\toprule
Initializer & First-frame & Per-frame \\
\midrule
Coarse-to-fine & \textbf{16.2} & \textbf{22.2} \\
One-step & 23.1 & 30.8 \\
\bottomrule
\end{tabular}
\end{minipage}
\end{table}

\begin{table}[t]
\centering
\caption{\textbf{Anchor prediction errors across seen and unseen marker configurations.} \texttt{config.a}--\texttt{config.c} are real configurations included during training and distinct from the baseline-specific layouts. \texttt{config.d} is held out from training and evaluated directly or after fine-tuning with either an estimated proxy configuration or the exact configuration.}
\label{tab:table2}
\resizebox{0.47\textwidth}{!}{
\tiny
\begin{tabular}{l|l|*{3}{c}}
\toprule
\multirow{2}{*}{Configuration} & \multirow{2}{*}{Setting} & \multicolumn{3}{c}{Data w/ occlusion and outliers} \\
        & & MPJPE & Body MPJPE & Hand MPJPE \\
\midrule
\texttt{config.a} & Seen & 10.19 & 10.31 & 10.11 \\
\texttt{config.b} & Seen & 9.01 & 7.66 & 10.00 \\
\texttt{config.c} & Seen & 9.56 & 10.17 & 9.11 \\
\midrule
\multirow{3}{*}{\texttt{config.d}} & Unseen (direct) & 12.93 & 10.52 & 14.70 \\
& Proxy-config fine-tuning & 10.28 & 7.76 & 12.14 \\
& Exact-config fine-tuning & 7.57 & 7.49 & 7.63 \\
\bottomrule
\end{tabular}
}
\end{table}

\noindent\textbf{Generalization to Unseen Marker Configurations.}
\textcolor{black}{To evaluate generalization to an unseen marker configuration, we test the trained model on three real configurations seen during training (\texttt{config.a}--\texttt{config.c}) and on a held-out \texttt{config.d}. The seen configurations are distinct from the marker layouts used by the baselines.} Because the solver operates on fixed anchor-to-body correspondences, we report anchor prediction errors that directly reflect initialization and trajectory quality. We first evaluate \texttt{config.d} directly without adaptation. We then estimate a proxy configuration, $\texttt{config.d}^{*}$, by voting over fitted SMPL-H vertices and use it for proxy-configuration fine-tuning; fine-tuning with the exact configuration provides a reference upper bound. As shown in \cref{tab:table2}, proxy-configuration fine-tuning improves accuracy over direct evaluation. Its smaller gain for hand joints is likely due to dense hand vertices making voting less reliable; for higher precision, hand vertices can instead be manually selected on the fitted SMPL-H mesh. This option is unavailable to configuration-specific baselines.

\section{HKMALA-Motion Dataset}
\label{sec:dataset}

\noindent\textbf{Raw Motion Collection.}
The collection contains 134 real traditional Chinese martial-arts
sequences captured between September 26, 2013, and March 8, 2022,
averaging approximately 8,500 frames and totaling about 180 minutes.
The metadata covers 22 normalized martial-arts style labels,
including Hung Kuen, Pakmei, Ving Tsun, Choy Lei Fut, Fujian White
Crane, Praying Mantis, Eagle Claw, Taiji, Bagua Zhang, and Tong Bei.
The recordings lack marker-layout metadata, marker identities,
and marker-to-body correspondences.

\noindent\textbf{Dataset Construction.}
We apply \modelname directly to each noisy and unordered marker sequence
without layout-specific annotations, converting the observations into a common
SMPL-H representation of pose, global translation, and shape through predicted
anchor trajectories and the differentiable solver. We retain the available
metadata, including capture date, martial-arts style, form name when available,
frame rate, and weapon-use flag. For weapon-based routines, the reconstruction
captures body motion only, as weapon geometry and trajectories lie outside the
SMPL-H model; the resulting parameters are therefore model-derived rather than
manually verified ground-truth annotations.

\noindent\textbf{Potential Applications.}
The common parametric representation supports character animation and retargeting, martial-arts motion retrieval and generation, and data-driven analysis and digital preservation of traditional movement. Its long, rapid, and stylistically diverse motions also facilitate long-horizon modeling, while the raw-to-parametric pipeline provides a basis for future joint human--weapon reconstruction.

\section{Conclusion}

In summary, \modelname recovers temporally coherent SMPL-H motion from unconstrained, noisy, and unordered markers without predefined layouts. By combining geometric proxy anchors, sliding-window tracking, and an end-to-end differentiable Gauss--Newton solver that learns input-dependent observation confidence, smoothness, and prior weights, a single model generalizes across diverse marker configurations and achieves competitive or superior accuracy to configuration-specific baselines. Despite these strong results, accuracy decreases on unseen, out-of-distribution configurations, and proxy-configuration voting remains unreliable in dense hand regions, occasionally requiring manual vertex selection. Extending the current SMPL-H implementation to another parametric model requires model-specific anchors, solver residuals and Jacobians, and retraining, although the initializer and trajectory-prediction designs remain reusable. Future work will automate configuration estimation for the hands and face and map recovered motion back to the original marker observations after labeling and outlier rejection, producing cleaned marker trajectories for arbitrary marker setups.

\section{Acknowledgment}
\label{sec:acknowledgment}

We thank International Guoshu Association Limited and the Institute of Chinese Martial Studies Limited for providing access to the raw martial-arts motion-capture recordings used to construct the HKMALA-Motion dataset. This work is funded by the Research Center for Industries of the Future (RCIF) at Westlake University and the Westlake Education Foundation.

\clearpage

\newpage

\clearpage

\bibliographystyle{ACM-Reference-Format}
\bibliography{references}

@String(CVPR 	=	{{Proc. IEEE Conf. on Computer Vision and Pattern Recognition (CVPR)}})

@String(ICCV 	=	{{International Conference on Computer Vision ({ICCV})}})

@String(ECCV 	=	{{European Conference on Computer Vision (ECCV)}})

@String(TOG 	=	{{Transactions on Graphics (TOG)}})

@inproceedings{cai2026omnifit,
  title={OmniFit: Multi-modal 3D Body Fitting via Scale-agnostic Dense Landmark Prediction},
  author={Cai, Zeyu and Xiu, Yuliang and Wang, Renke and Shao, Zhijing and Li, Xiaoben and Yu, Siyuan and Xu, Chao and Liu, Yang and Sun, Baigui and Yang, Jian and others},
  booktitle={European Conference on Computer Vision (ECCV)},
  note={Accepted for publication},
  year={2026}
}

@conference{AMASS:ICCV:2019,
  title = {{AMASS}: Archive of Motion Capture as Surface Shapes},
  author = {Mahmood, Naureen and Ghorbani, Nima and Troje, Nikolaus F. and Pons-Moll, Gerard and Black, Michael J.},
  booktitle = ICCV,
  pages = {5442--5451},
  month = oct,
  year = {2019},
  month_numeric = {10}
}

@inproceedings{li2026etch,
  title={ETCH-X: Robustify Expressive Body Fitting to Clothed Humans with Composable Datasets},
  author={Li, Xiaoben and Wu, Jingyi and Cai, Zeyu and Yu, Siyuan and Li, Boqian and Xiu, Yuliang},
  booktitle={European Conference on Computer Vision (ECCV)},
  note={Accepted for publication},
  year={2026}
}

@incollection{kirk2004skeletal,
  title={Skeletal parameter estimation from optical motion capture data},
  author={Kirk, Adam and O'Brien, James F and Forsyth, David A},
  booktitle={ACM SIGGRAPH 2004 Sketches},
  pages={29},
  year={2004}
}

@article{burke2016estimating,
  title={Estimating missing marker positions using low dimensional Kalman smoothing},
  author={Burke, Michael and Lasenby, Joan},
  journal={Journal of biomechanics},
  volume={49},
  number={9},
  pages={1854--1858},
  year={2016},
  publisher={Elsevier}
}

@article{feng2014exploiting,
  title={Exploiting temporal stability and low-rank structure for motion capture data refinement},
  author={Feng, Yinfu and Xiao, Jun and Zhuang, Yueting and Yang, Xiaosong and Zhang, Jian J and Song, Rong},
  journal={Information Sciences},
  volume={277},
  pages={777--793},
  year={2014},
  publisher={Elsevier}
}

@article{aristidou2018self,
  title={Self-similarity analysis for motion capture cleaning},
  author={Aristidou, Andreas and Cohen-Or, Daniel and Hodgins, Jessica K and Shamir, Ariel},
  journal={Computer Graphics Forum},
  volume={37},
  number={2},
  pages={297--309},
  year={2018},
  publisher={Wiley},
  doi={10.1111/cgf.13362}
}

@article{tits2018robust,
  title={Robust and automatic motion-capture data recovery using soft skeleton constraints and model averaging},
  author={Tits, Micka{\"e}l and Tilmanne, Jo{\"e}lle and Dutoit, Thierry},
  journal={PloS one},
  volume={13},
  number={7},
  pages={e0199744},
  year={2018},
  publisher={Public Library of Science San Francisco, CA USA}
}

@article{holden2018robust,
  title={Robust solving of optical motion capture data by denoising},
  author={Holden, Daniel},
  journal={ACM Transactions on Graphics (TOG)},
  volume={37},
  number={4},
  pages={1--12},
  year={2018},
  publisher={ACM New York, NY, USA}
}

@inproceedings{li2026droid,
  title={DROID-SLAM in the Wild},
  author={Li, Moyang and Zhu, Zihan and Pollefeys, Marc and Barath, Daniel},
  booktitle={Proceedings of the IEEE/CVF Conference on Computer Vision and Pattern Recognition (CVPR)},
  pages={36498--36508},
  month=jun,
  year={2026}
}

@inproceedings{xu2022rnnpose,
  title={Rnnpose: Recurrent 6-dof object pose refinement with robust correspondence field estimation and pose optimization},
  author={Xu, Yan and Lin, Kwan-Yee and Zhang, Guofeng and Wang, Xiaogang and Li, Hongsheng},
  booktitle={Proceedings of the IEEE/CVF conference on computer vision and pattern recognition},
  pages={14880--14890},
  year={2022}
}

@inproceedings{iwase2021repose,
  title={Repose: Fast 6d object pose refinement via deep texture rendering},
  author={Iwase, Shun and Liu, Xingyu and Khirodkar, Rawal and Yokota, Rio and Kitani, Kris M},
  booktitle={Proceedings of the IEEE/CVF International Conference on Computer Vision},
  pages={3303--3312},
  year={2021}
}

@inproceedings{li2025megasam,
  title={Megasam: Accurate, fast and robust structure and motion from casual dynamic videos},
  author={Li, Zhengqi and Tucker, Richard and Cole, Forrester and Wang, Qianqian and Jin, Linyi and Ye, Vickie and Kanazawa, Angjoo and Holynski, Aleksander and Snavely, Noah},
  booktitle={Proceedings of the IEEE/CVF Conference on Computer Vision and Pattern Recognition},
  pages={10486--10496},
  year={2025}
}

@inproceedings{pavlakos2019expressive,
  title={Expressive body capture: 3d hands, face, and body from a single image},
  author={Pavlakos, Georgios and Choutas, Vasileios and Ghorbani, Nima and Bolkart, Timo and Osman, Ahmed AA and Tzionas, Dimitrios and Black, Michael J},
  booktitle={Proceedings of the IEEE/CVF conference on computer vision and pattern recognition},
  pages={10975--10985},
  year={2019}
}

@article{ferguson2025mhr,
  title={Mhr: Momentum human rig},
  author={Ferguson, Aaron and Osman, Ahmed AA and Bescos, Berta and Stoll, Carsten and Twigg, Chris and Lassner, Christoph and Otte, David and Vignola, Eric and Prada, Fabian and Bogo, Federica and others},
  journal={arXiv preprint arXiv:2511.15586},
  year={2025}
}

@inproceedings{wu2024point,
  title={Point transformer v3: Simpler faster stronger},
  author={Wu, Xiaoyang and Jiang, Li and Wang, Peng-Shuai and Liu, Zhijian and Liu, Xihui and Qiao, Yu and Ouyang, Wanli and He, Tong and Zhao, Hengshuang},
  booktitle={Proceedings of the IEEE/CVF conference on computer vision and pattern recognition},
  pages={4840--4851},
  year={2024}
}

@article{chen2021mocap,
  title={Mocap-solver: A neural solver for optical motion capture data},
  author={Chen, Kang and Wang, Yupan and Zhang, Song-Hai and Xu, Sen-Zhe and Zhang, Weidong and Hu, Shi-Min},
  journal={ACM Transactions on Graphics (TOG)},
  volume={40},
  number={4},
  pages={1--11},
  year={2021},
  publisher={ACM New York, NY, USA}
}

@inproceedings{ghorbani2021soma,
  title={Soma: Solving optical marker-based mocap automatically},
  author={Ghorbani, Nima and Black, Michael J},
  booktitle={Proceedings of the IEEE/CVF International Conference on Computer Vision},
  pages={11117--11126},
  year={2021}
}

@inproceedings{pan2023locality,
  title={A locality-based neural solver for optical motion capture},
  author={Pan, Xiaoyu and Zheng, Bowen and Jiang, Xinwei and Xu, Guanglong and Gu, Xianli and Li, Jingxiang and Kou, Qilong and Wang, He and Shao, Tianjia and Zhou, Kun and others},
  booktitle={SIGGRAPH Asia 2023 conference papers},
  pages={1--11},
  year={2023}
}

@inproceedings{pan2024romo,
  title={RoMo: A Robust Solver for Full-body Unlabeled Optical Motion Capture},
  author={Pan, Xiaoyu and Zheng, Bowen and Jiang, Xinwei and Zeng, Zijiao and Kou, Qilong and Wang, He and Jin, Xiaogang},
  booktitle={SIGGRAPH Asia 2024 Conference Papers},
  pages={1--11},
  year={2024}
}

@article{kim2024damo,
  title={Damo: A deep solver for arbitrary marker configuration in optical motion capture},
  author={Kim, KyeongMin and Seo, SeungWon and Han, DongHeun and Kang, HyeongYeop},
  journal={ACM Transactions on Graphics},
  volume={44},
  number={1},
  pages={1--14},
  year={2024},
  publisher={ACM New York, NY}
}

@inproceedings{qian2025openmocap,
  title={OpenMoCap: Rethinking Optical Motion Capture under Real-world Occlusion},
  author={Qian, Chen and Li, Danyang and Yu, Xinran and Yang, Zheng and Ma, Qiang},
  booktitle={Proceedings of the 33rd ACM International Conference on Multimedia},
  pages={7529--7537},
  year={2025}
}

@inproceedings{karaev2024cotracker,
  title={Cotracker: It is better to track together},
  author={Karaev, Nikita and Rocco, Ignacio and Graham, Benjamin and Neverova, Natalia and Vedaldi, Andrea and Rupprecht, Christian},
  booktitle={European conference on computer vision},
  pages={18--35},
  year={2024},
  organization={Springer}
}

@inproceedings{xiao2024spatialtracker,
  title={Spatialtracker: Tracking any 2d pixels in 3d space},
  author={Xiao, Yuxi and Wang, Qianqian and Zhang, Shangzhan and Xue, Nan and Peng, Sida and Shen, Yujun and Zhou, Xiaowei},
  booktitle={Proceedings of the IEEE/CVF Conference on Computer Vision and Pattern Recognition},
  pages={20406--20417},
  year={2024}
}

@inproceedings{lu2025dposer,
  title={DPoser-X: Diffusion model as robust 3D whole-body human pose prior},
  author={Lu, Junzhe and Lin, Jing and Dou, Hongkun and Zeng, Ailing and Deng, Yue and Liu, Xian and Cai, Zhongang and Yang, Lei and Zhang, Yulun and Wang, Haoqian and others},
  booktitle={Proceedings of the IEEE/CVF International Conference on Computer Vision},
  pages={9988--9997},
  year={2025}
}

@inproceedings{patel2025camerahmr,
  title={Camerahmr: Aligning people with perspective},
  author={Patel, Priyanka and Black, Michael J},
  booktitle={2025 International Conference on 3D Vision (3DV)},
  pages={1562--1571},
  year={2025},
  organization={IEEE}
}

@inproceedings{li2025etch,
  title={Etch: Generalizing body fitting to clothed humans via equivariant tightness},
  author={Li, Boqian and Feng, Haiwen and Cai, Zeyu and Black, Michael J and Xiu, Yuliang},
  booktitle={Proceedings of the IEEE/CVF International Conference on Computer Vision},
  pages={8264--8274},
  year={2025}
}

@inproceedings{teed2021droid,
  title={Droid-slam: Deep visual slam for monocular, stereo, and rgb-d cameras},
  author={Teed, Zachary and Deng, Jia},
  booktitle={Advances in Neural Information Processing Systems},
  volume={34},
  pages={16558--16569},
  year={2021}
}

@inproceedings{sarlin2021back,
  title={Back to the feature: Learning robust camera localization from pixels to pose},
  author={Sarlin, Paul-Edouard and Unagar, Ajaykumar and Larsson, Mans and Germain, Hugo and Toft, Carl and Larsson, Viktor and Pollefeys, Marc and Lepetit, Vincent and Hammarstrand, Lars and Kahl, Fredrik and others},
  booktitle={Proceedings of the IEEE/CVF conference on computer vision and pattern recognition},
  pages={3247--3257},
  year={2021}
}

@inproceedings{chen2022epro,
  title={Epro-pnp: Generalized end-to-end probabilistic perspective-n-points for monocular object pose estimation},
  author={Chen, Hansheng and Wang, Pichao and Wang, Fan and Tian, Wei and Xiong, Lu and Li, Hao},
  booktitle={Proceedings of the IEEE/CVF conference on computer vision and pattern recognition},
  pages={2781--2790},
  year={2022}
}

@inproceedings{wang2023deep,
  title={Deep active contours for real-time 6-DoF object tracking},
  author={Wang, Long and Yan, Shen and Zhen, Jianan and Liu, Yu and Zhang, Maojun and Zhang, Guofeng and Zhou, Xiaowei},
  booktitle={Proceedings of the IEEE/CVF International Conference on Computer Vision},
  pages={14034--14044},
  year={2023}
}

@inproceedings{taheri2020grab,
  title={GRAB: A dataset of whole-body human grasping of objects},
  author={Taheri, Omid and Ghorbani, Nima and Black, Michael J and Tzionas, Dimitrios},
  booktitle={European conference on computer vision},
  pages={581--600},
  year={2020},
  organization={Springer}
}

@misc{cmu_mocap,
  title        = {CMU Graphics Lab Motion Capture Database},
  howpublished = {\url{http://mocap.cs.cmu.edu/}},
  year         = {2000},
  organization = {CMU Graphics Lab}
}

\newpage
\appendix

\addcontentsline{toc}{section}{Appendix} %

\section{Body-centric Canonical Space}
\label{Body Centric Canonical Space}
From the anatomical anchors at time $t$
\begin{equation}
\mathbf{J}_{t}^{\mathrm{ana}} = \{\mathbf{J}_{t}^{\mathrm{pelvis}}, \mathbf{J}_{t}^{\mathrm{neck}}, \mathbf{J}_{t}^{\mathrm{lhip}}, \mathbf{J}_{t}^{\mathrm{rhip}}, \mathbf{J}_{t}^{\mathrm{lshoulder}}, \mathbf{J}_{t}^{\mathrm{rshoulder}}\} \in \mathbb{R}^{6\times 3},
\end{equation}
where $\mathbf{J}_{t}^{\mathrm{ana}}$ can be obtained either from the anchor initialization module~\cref{sub:anchor_initialization} or by directly selecting the corresponding anchors from $\mathbf{A}_t$ according to the anatomical indices,
we construct a body-centric canonical space. The lateral (X) axis is computed from the averaged hip and shoulder vectors:
\begin{equation}
\begin{split}
\mathbf{X} &= \frac{\mathbf{v}_\text{hip} + \mathbf{v}_\text{shoulder}}{\|\mathbf{v}_\text{hip} + \mathbf{v}_\text{shoulder}\|} \\
    \mathbf{v}_\text{hip} = \mathbf{J}_{t}^{\mathrm{lhip}} - \mathbf{J}_{t}^{\mathrm{rhip}}, &\quad
\mathbf{v}_\text{shoulder} = \mathbf{J}_{t}^{\mathrm{lshoulder}} - \mathbf{J}_{t}^{\mathrm{rshoulder}}.
\end{split}
\end{equation}
The longitudinal (Y) axis is defined from pelvis to neck:
\begin{equation}
    \mathbf{Y} = \frac{\mathbf{J}_{t}^{\mathrm{neck}} - \mathbf{J}_{t}^{\mathrm{pelvis}}}{\|\mathbf{J}_{t}^{\mathrm{neck}} - \mathbf{J}_{t}^{\mathrm{pelvis}}\|}.
\end{equation}
The vertical (Z) axis is obtained via cross product, followed by re-orthogonalization of Y:
\begin{equation}
    \mathbf{Z} = \frac{\mathbf{X} \times \mathbf{Y}}{\|\mathbf{X} \times \mathbf{Y}\|}, \quad
\mathbf{Y} := \mathbf{Z} \times \mathbf{X}.
\end{equation}
The canonical transformation of $\mathbf{J}_{t}^{\mathrm{ana}}$ is defined
\begin{equation}
\begin{split}
x^\text{cano} = & \mathcal{T}_{t}(x), \quad \text{where } \mathcal{T}_{t}(x) = \mathbf{R}(x - \mathbf{T}). \\
\mathbf{R} = & [\mathbf{X}, \mathbf{Y}, \mathbf{Z}] \in \mathbb{R}^{3 \times 3}, \quad \mathbf{T} = \mathbf{J}_{t}^{\mathrm{pelvis}}.
\end{split}
\end{equation}
Through the above transformation, both marker observations and anchor positions can be mapped into the body-centric canonical space defined by $\mathbf{J}_{t}^{\mathrm{ana}}$.

\section{Parametric Model: SMPL-H}
\label{Parametric Model: SMPL-H}
We use SMPL-H~\cite{pavlakos2019expressive} as the underlying parametric human body model. SMPL-H extends the SMPL body model with articulated hands and represents the human body as a skinned triangular mesh driven by pose and shape parameters. In our formulation, the pose parameter at frame $i$ is denoted as $\theta_i \in \mathbb{R}^{156}$, corresponding to the axis-angle rotations of 52 joints, and the global translation is denoted as $\tau_i \in \mathbb{R}^{3}$. The shape parameter $\beta \in \mathbb{R}^{10}$ is shared across the sequence.
Given pose $\theta_i$ and shape $\beta$, SMPL-H first constructs a posed and shaped template mesh from a canonical rest template. Specifically, the rest template is deformed by shape-dependent blend shapes and pose-dependent corrective blend shapes:
\begin{equation}
\mathbf{T}(\theta_i,\beta)
=
\bar{\mathbf{T}}
+
B_S(\beta)
+
B_P(\theta_i),
\end{equation}
where $\bar{\mathbf{T}}$ is the template mesh, $B_S(\beta)$ denotes the shape blend-shape function, and $B_P(\theta_i)$ denotes the pose-dependent corrective blend-shape function.

The joint locations are regressed from the shaped body template using a learned joint regressor:
\begin{equation}
\mathbf{J}(\beta) = \mathcal{R}_{J}\left(\bar{\mathbf{T}} + B_S(\beta)\right),
\end{equation}
where $\mathcal{R}_{J}$ denotes the SMPL-H joint regression operator. Given the joint locations and pose rotations, forward kinematics computes the global rigid transformation of each joint along the kinematic tree. The mesh vertices are then transformed using Linear Blend Skinning (LBS):
\begin{equation}
\mathbf{V}(\theta_i,\beta)
=
\mathcal{W}\left(
\mathbf{T}(\theta_i,\beta),
\mathbf{J}(\beta),
\theta_i,
\mathcal{W}_{\mathrm{skin}}
\right),
\end{equation}
where $\mathcal{W}(\cdot)$ denotes the LBS function and $\mathcal{W}_{\mathrm{skin}}$ denotes the skinning weights. Finally, the global translation is applied to obtain vertices in the world coordinate system:
\begin{equation}
\mathbf{V}_i = \mathbf{V}(\theta_i,\beta) + \tau_i.
\end{equation}

In our method, we use both SMPL-H joints and selected surface vertices as geometric anchors. Let $\mathcal{I}_{S}$ denote the index set of the selected surface vertices. The anchor function used in the main paper is defined as
\begin{equation}
\hat{\mathbf{A}}(\theta_i,\tau_i,\beta)
=
\left\{
\hat{\mathbf{J}}_i,
\hat{\mathbf{S}}_i
\right\},
\end{equation}
where
\begin{equation}
\hat{\mathbf{J}}_i = \mathbf{J}_{\mathrm{posed}}(\theta_i,\beta) + \tau_i,
\qquad
\hat{\mathbf{S}}_i = \mathbf{V}_i[\mathcal{I}_{S}].
\end{equation}
Here, $\hat{\mathbf{J}}_i \in \mathbb{R}^{52\times3}$ contains the posed SMPL-H body and hand joints, and $\hat{\mathbf{S}}_i \in \mathbb{R}^{61\times3}$ contains the selected surface anchors. Therefore, the full anchor set contains $K=113$ points.

This formulation provides a differentiable mapping from the SMPL-H parameters $(\theta_i,\tau_i,\beta)$ to the geometric anchors $\hat{\mathbf{A}}(\theta_i,\tau_i,\beta)$ used by the observation and smoothness terms in our solver.

\section{Jacobian of SMPL-H Anchors}
\label{Jacobian of SMPL-H Anchors}

In the differentiable Gauss--Newton solver, we compute the Jacobian of the SMPL-H anchor function with respect to the optimization variables. For each window, the unknowns are
\begin{equation}
\mathbf{x} = [\theta_{t:t+W-1}, \tau_{t:t+W-1}, \beta],
\end{equation}
where $\theta_i$ denotes the axis-angle pose at frame $i$, $\tau_i$ is the global translation, and $\beta$ is the sequence-level shape parameter. For each frame $i$, the anchor Jacobian is written as
\begin{equation}
\frac{\partial \hat{\mathbf{A}}(\theta_i,\tau_i,\beta)}
{\partial \mathbf{x}}
=
\left[
\frac{\partial \hat{\mathbf{A}}_i}{\partial \theta_i},
\frac{\partial \hat{\mathbf{A}}_i}{\partial \tau_i},
\frac{\partial \hat{\mathbf{A}}_i}{\partial \beta}
\right],
\end{equation}
where $\hat{\mathbf{A}}_i$ is shorthand for $\hat{\mathbf{A}}(\theta_i,\tau_i,\beta)$.

For a surface anchor attached to vertex $v_j$, its posed position under Linear Blend Skinning can be written as
\begin{equation}
\hat{\mathbf{a}}_{i,j}
=
\Pi\left(
\sum_{b \in \mathcal{B}_j}
\omega_{j,b}
\mathbf{G}'_{b}(\theta_i,\beta)
\begin{bmatrix}
\bar{\mathbf{v}}_{j}(\beta)\\
1
\end{bmatrix}
\right)
+
\tau_i,
\end{equation}
where $\Pi(\cdot)$ extracts the first three coordinates of a homogeneous vector, $\mathcal{B}_j$ denotes the set of bones that influence vertex $v_j$, $\omega_{j,b}$ is the corresponding skinning weight, $\bar{\mathbf{v}}_j(\beta)$ is the shaped rest-space vertex, and $\mathbf{G}'_{b}$ is the relative bone transformation used by LBS. The same formulation applies to joint anchors by replacing the surface vertex with the corresponding joint location.

The global transformation of a bone is obtained by multiplying local transformations along its kinematic chain:
\begin{equation}
\mathbf{G}_{b}(\theta_i,\beta)
=
\prod_{q \in \mathcal{A}(b)}
\begin{bmatrix}
\exp(\theta_{i,q}) & \bar{\mathbf{t}}_{q}(\beta) \\
\mathbf{0}^{\top} & 1
\end{bmatrix},
\end{equation}
where $\mathcal{A}(b)$ denotes the ordered set of joints along the chain from the root to bone $b$, $\exp(\theta_{i,q})$ converts the axis-angle vector into a rotation matrix, and $\bar{\mathbf{t}}_{q}(\beta)$ is the rest-pose offset between joint $q$ and its parent. The transformation used by LBS is the relative transformation with respect to the rest pose:
\begin{equation}
\mathbf{G}'_{b}(\theta_i,\beta)
=
\mathbf{G}_{b}(\theta_i,\beta)
\begin{bmatrix}
\mathbf{I}_{3} & -\bar{\mathbf{j}}_{b}(\beta) \\
\mathbf{0}^{\top} & 1
\end{bmatrix},
\end{equation}
where $\bar{\mathbf{j}}_{b}(\beta)$ is the rest-space joint location associated with bone $b$. Equivalently,
\begin{equation}
\mathbf{G}'_{b}(\theta_i,\beta)
\begin{bmatrix}
\bar{\mathbf{v}}_j(\beta) \\
1
\end{bmatrix}
=
\mathbf{G}_{b}(\theta_i,\beta)
\begin{bmatrix}
\bar{\mathbf{v}}_j(\beta)-\bar{\mathbf{j}}_{b}(\beta) \\
1
\end{bmatrix}.
\end{equation}
Thus, $\mathbf{G}'_{b}$ first maps a rest-space point into the local coordinate system of bone $b$ by subtracting $\bar{\mathbf{j}}_{b}(\beta)$, and then transforms it to the posed configuration using $\mathbf{G}_{b}(\theta_i,\beta)$.

In Gauss--Newton optimization, we linearize the anchor function with respect to the parameter increments. For the pose update, we use
\begin{equation}
\exp(\theta_{i,q}+\Delta\theta_{i,q})
\approx
\exp(\theta_{i,q})\exp(\Delta\theta_{i,q}),
\end{equation}
and compute the derivative with respect to the small update $\Delta\theta_{i,q}$. Therefore, for the $r$-th component of $\Delta\theta_{i,q}$, the derivative of a chain transformation is obtained by replacing the local rotation at joint $q$ with the derivative of the exponential map:
\begin{equation}
\begin{split}
\frac{\partial \mathbf{G}_{b}}
{\partial \Delta\theta_{i,q}^{r}}
=
&\left(
\prod_{p \in \mathcal{A}(b),\,p<q}
\begin{bmatrix}
\exp(\theta_{i,p}) & \bar{\mathbf{t}}_{p}(\beta) \\
\mathbf{0}^{\top} & 1
\end{bmatrix}
\right)
\\
&\cdot
\begin{bmatrix}
\exp(\theta_{i,q})
\left.
\frac{\partial \exp(\Delta\theta)}
{\partial \Delta\theta^{r}}
\right|_{\Delta\theta=\mathbf{0}}
&
\mathbf{0}
\\
\mathbf{0}^{\top} & 0
\end{bmatrix}
\\
&\cdot
\left(
\prod_{p \in \mathcal{A}(b),\,p>q}
\begin{bmatrix}
\exp(\theta_{i,p}) & \bar{\mathbf{t}}_{p}(\beta) \\
\mathbf{0}^{\top} & 1
\end{bmatrix}
\right).
\end{split}
\end{equation}
Using the chain rule, the derivative of a surface anchor with respect to the pose increment is
\begin{equation}
\frac{\partial \hat{\mathbf{a}}_{i,j}}
{\partial \Delta\theta_{i,q}^{r}}
=
\Pi\left[
\sum_{b \in \mathcal{B}_j}
\omega_{j,b}
\frac{\partial \mathbf{G}'_{b}(\theta_i,\beta)}
{\partial \Delta\theta_{i,q}^{r}}
\begin{bmatrix}
\bar{\mathbf{v}}_{j}(\beta) \\
1
\end{bmatrix}
\right],
\end{equation}
where the derivative is non-zero only when joint $q$ lies on the kinematic chain of an influencing bone $b$. Since the rest-space correction 
$\begin{bmatrix}
\mathbf{I}_{3} & -\bar{\mathbf{j}}_{b}(\beta) \\
\mathbf{0}^{\top} & 1
\end{bmatrix}$
does not depend on the pose, we have
\begin{equation}
\frac{\partial \mathbf{G}'_{b}(\theta_i,\beta)}
{\partial \Delta\theta_{i,q}^{r}}
=
\frac{\partial \mathbf{G}_{b}(\theta_i,\beta)}
{\partial \Delta\theta_{i,q}^{r}}
\begin{bmatrix}
\mathbf{I}_{3} & -\bar{\mathbf{j}}_{b}(\beta) \\
\mathbf{0}^{\top} & 1
\end{bmatrix}.
\end{equation}
Therefore, each anchor depends not only on the bones that directly skin the corresponding vertex, but also on the ancestor joints along the kinematic chains of these bones.

The derivative with respect to global translation is straightforward, since $\tau_i$ is added to all anchors:
\begin{equation}
\frac{\partial \hat{\mathbf{A}}_i}
{\partial \tau_i}
=
\mathbf{1}_{K}\otimes\mathbf{I}_{3},
\end{equation}
where $\mathbf{1}_{K}$ is a $K$-dimensional vector of ones and $\mathbf{I}_{3}$ is the $3\times3$ identity matrix.

For the shape parameter, the derivative contains two parts. First, $\beta$ changes the shaped rest-space anchors through the SMPL-H shape blend-shape basis. Second, $\beta$ changes the rest-space joint locations, and therefore the bone offsets and the relative transformation used in LBS. For a surface anchor, the shape derivative is written as
\begin{equation}
\begin{split}
\frac{\partial \hat{\mathbf{a}}_{i,j}}
{\partial \beta_m}
=
\Pi\biggl[
\sum_{b \in \mathcal{B}_j}
\omega_{j,b}
\Bigg(
&
\frac{\partial \mathbf{G}'_{b}(\theta_i,\beta)}
{\partial \beta_m}
\begin{bmatrix}
\bar{\mathbf{v}}_{j}(\beta) \\
1
\end{bmatrix}
\\
&+
\mathbf{G}'_{b}(\theta_i,\beta)
\begin{bmatrix}
\frac{\partial \bar{\mathbf{v}}_{j}(\beta)}
{\partial \beta_m} \\
0
\end{bmatrix}
\Bigg)
\biggr].
\end{split}
\end{equation}
Here, $\frac{\partial \bar{\mathbf{v}}_{j}(\beta)}{\partial \beta_m}$ is directly given by the $m$-th SMPL-H shape basis. The term $\frac{\partial \mathbf{G}'_{b}}{\partial \beta_m}$ comes from the dependence of the kinematic chain on the shaped rest-pose joint locations. Specifically, since each rest-pose offset $\bar{\mathbf{t}}_{p}(\beta)$ is determined by the shaped skeleton, the derivative of the global bone transformation with respect to $\beta_m$ is written as
\begin{equation}
\frac{\partial \mathbf{G}_{b}}
{\partial \beta_m}
=
\prod_{p \in \mathcal{A}(b)}
\begin{bmatrix}
\exp(\theta_{i,p}) & \frac{\partial \bar{\mathbf{t}}_{p}(\beta)}{\partial \beta_m} \\
\mathbf{0}^{\top} & 1
\end{bmatrix}.
\end{equation}
Since the relative LBS transformation additionally depends on the rest-space joint location $\bar{\mathbf{j}}_b(\beta)$, its shape derivative is
\begin{equation}
\begin{split}
\frac{\partial \mathbf{G}'_{b}}
{\partial \beta_m}
=
&
\frac{\partial \mathbf{G}_{b}}
{\partial \beta_m}
\begin{bmatrix}
\mathbf{I}_{3} & -\bar{\mathbf{j}}_{b}(\beta)  \\
\mathbf{0}^{\top} & 1
\end{bmatrix}
\\
&+
\mathbf{G}_{b}(\theta_i,\beta)
\begin{bmatrix}
\mathbf{0}_{3\times3} &
-\frac{\partial \bar{\mathbf{j}}_{b}(\beta)}{\partial \beta_m}  \\
\mathbf{0}^{\top} & 0
\end{bmatrix}.
\end{split}
\end{equation}

The rest-space joint locations are obtained from the shaped mesh using the SMPL-H joint regressor:
\begin{equation}
\bar{\mathbf{j}}_{q}(\beta)
=
\sum_{j}
\mathcal{R}_{qj}\bar{\mathbf{v}}_{j}(\beta),
\qquad
\frac{\partial \bar{\mathbf{j}}_{q}(\beta)}
{\partial \beta_m}
=
\sum_{j}
\mathcal{R}_{qj}
\frac{\partial \bar{\mathbf{v}}_{j}(\beta)}
{\partial \beta_m}.
\end{equation}
The rest-pose offset between joint $q$ and its parent $\pi(q)$ is
\begin{equation}
\bar{\mathbf{t}}_{q}(\beta)
=
\bar{\mathbf{j}}_{q}(\beta)
-
\bar{\mathbf{j}}_{\pi(q)}(\beta),
\qquad
\frac{\partial \bar{\mathbf{t}}_{q}(\beta)}
{\partial \beta_m}
=
\frac{\partial \bar{\mathbf{j}}_{q}(\beta)}
{\partial \beta_m}
-
\frac{\partial \bar{\mathbf{j}}_{\pi(q)}(\beta)}
{\partial \beta_m}.
\end{equation}
Thus, both $\frac{\partial \bar{\mathbf{j}}_{q}(\beta)}{\partial \beta_m}$ and $\frac{\partial \bar{\mathbf{t}}_{q}(\beta)}{\partial \beta_m}$ are computed from the SMPL-H shape blend-shape basis and joint regressor.

For efficiency, we implement the Jacobian computation in CUDA. Instead of relying on automatic differentiation through the full SMPL-H forward pass, our implementation explicitly computes the derivatives of the selected SMPL-H anchors with respect to pose, translation, and shape. The computation is parallelized over frames, anchors, and Jacobian entries, allowing the required Jacobians and the corresponding Gauss--Newton system terms to be evaluated efficiently.

\end{document}